\documentclass{article}

\title{In Search of Robust Measures of Generalization}

\usepackage[nonatbib,final]{neurips_2020}
\usepackage{bbm}

\pdfoutput=1
\usepackage{amsmath,amsthm,wrapfig}
	\usepackage[utf8]{inputenc} 
\usepackage{amsmath, amssymb,bm, cases, mathtools, thmtools}
\usepackage{verbatim}
\usepackage{graphicx}\graphicspath{{figures/}}
\usepackage{multicol}
\usepackage{tabularx}
\usepackage[usenames,dvipsnames]{xcolor}
\usepackage{mathrsfs} 
\usepackage{caption}
\usepackage{algorithm}
\usepackage{algorithmicx}
\usepackage[noend]{algpseudocode}

\usepackage[%
    minnames=1,maxnames=99,maxcitenames=3,
    style=numeric,%
    sortcites=true,
    doi=false,url=false,
    giveninits=true,
    hyperref,natbib,backend=biber]{biblatex}
\renewbibmacro{in:}{%
  \ifentrytype{article}{}{\printtext{\bibstring{in}\intitlepunct}}}

\bibliography{biblio.bib}

\usepackage[T1]{fontenc}
\usepackage{inconsolata}

\usepackage{listings} %

\usepackage{enumitem}
\usepackage{booktabs}       %
\usepackage{nicefrac}       %

\usepackage{lineno}
\usepackage{bbm}

\usepackage{caption}
\usepackage{subcaption}

\usepackage[colorlinks,citecolor=blue,urlcolor=blue,linkcolor=RawSienna]{hyperref}
\usepackage{datetime}

\DeclareMathAlphabet\EuRoman{U}{eur}{m}{n}
\SetMathAlphabet\EuRoman{bold}{U}{eur}{b}{n}

\usepackage[capitalize]{cleveref}

\crefname{lemma}{Lemma}{Lemmas}
\crefname{corollary}{Corollary}{Corollaries}
\crefname{theorem}{Theorem}{Theorems}

\makeatletter
\let\reftagform@=\tagform@
\def\tagform@#1{\maketag@@@{\ignorespaces\textcolor{gray}{(#1)}\unskip\@@italiccorr}}
\renewcommand{\eqref}[1]{\textup{\reftagform@{\ref{#1}}}}
\makeatother

\numberwithin{theorem}{section}

\usepackage{xparse}
\usepackage{xstring}

\def\[#1\]{\begin{align}#1\end{align}}
\def\*[#1\]{\begin{align*}#1\end{align*}}

\newcommand{\defas}{\vcentcolon=}  %

\newcommand{\lossweight}{\chi}
\newcommand{\losstransform}{\kappa}
\newcommand{\neff}{n_{\text{eff}}}

\newcommand{\env}{\mathcal{E}}

\newcommand{\ip}[2]{\langle #1,#2\rangle}

\newcommand{\Dist}{\mathcal D}
\newcommand\optparen[1]{\ifthenelse{\equal{#1}{}}{}{(#1)}}
\newcommand{\RiskChar}{R}
\newcommand{\Risk}[2]{\RiskChar_{#1}\optparen{#2}}
\newcommand{\EmpRisk}[2]{\hat \RiskChar_{#1}\optparen{#2}}

\newcommand{\Reals}{\mathbb{R}}

\DeclareMathOperator{\sign}{\text{sgn}}

\DeclareMathOperator*{\newlim}{\mathrm{lim}\vphantom{\mathrm{infsup}}}
\DeclareMathOperator*{\newmin}{\mathrm{min}\vphantom{\mathrm{infsup}}}
\DeclareMathOperator*{\newmax}{\mathrm{max}\vphantom{\mathrm{infsup}}}
\DeclareMathOperator*{\newinf}{\mathrm{inf}\vphantom{\mathrm{infsup}}}
\DeclareMathOperator*{\newsup}{\mathrm{sup}\vphantom{\mathrm{infsup}}}
\renewcommand{\lim}{\newlim}
\renewcommand{\min}{\newmin}
\renewcommand{\max}{\newmax}
\renewcommand{\inf}{\newinf}
\renewcommand{\sup}{\newsup}

\renewcommand{\Pr}{\mathbb{P}}
\def\EE{\mathbb{E}}

\newcommand{\defn}[1]{\textbf{#1}}

\newcommand{\abs}[1]{\lvert #1 \rvert}
\newcommand{\norm}[1]{\lVert #1 \rVert}

\newcommand{\loss}{\ell}

\newcommand{\lcrx}[4][{-1}]{
	\IfEq{#1}{-1}{\left #2 {{{{#3}}}} \right #4}{
   	\IfEq{#1}{0}{#1 {{{{#2}}}} #3}{
	\IfEq{#1}{1}{\bigl #2 {{{{#3}}}} \bigr #4}{
	\IfEq{#1}{2}{\Bigl #2 {{{{#3}}}} \Bigr #4}{
	\IfEq{#1}{3}{\biggl #2 {{{{#3}}}} \biggr #4}{
	\IfEq{#1}{4}{\Biggl #2 {{{{#3}}}} \Biggr #4}{
    \GenericWarning{"4th argument to lcrx must be -1, 0, 1, 2, 3, or 4"}
    }}}}}}}

\usepackage[hide=true,setmargin=true,marginparwidth=1.25in]{marginalia}
\usepackage{xspace}

\usepackage{grffile}
\usepackage{wrapfig}
\usepackage{minitoc} %

\newcommand{\GM}{C}
\newcommand{\GenError}{G}
\newcommand{\MSE}[2]{\mathrm{MSE}(#1,#2)}
\newcommand{\SignE}[2]{\mathrm{SE}(#1,#2)}
\renewcommand{\defn}[1]{\emph{#1}}

\newcommand{\tildeO}{\smash{\tilde O}}
\newcommand{\ww}{\bm{w}}
\newcommand{\xx}{\bm{x}}

\usepackage{mathtools} %

\definecolor{limegreen}{rgb}{0.2,0.8,0.2}
\definecolor{magenta}{rgb}{1.0,0.0,1.0}
\definecolor{orange}{rgb}{1.0,0.5,0.0}

\usepackage{caption}
\usepackage{subcaption}

\begin{document}
\doparttoc %
\faketableofcontents %

\author{%
Gintare Karolina Dziugaite$^{1}$\thanks{Equal contribution. Correspondence to: \texttt{\{karolina.dziugaite, alexandre.drouin\}@elementai.com}}~,
Alexandre Drouin$^{1*}$,
Brady Neal$^{1, 2, 3}$,
Nitarshan Rajkumar$^{2, 3}$,
\\[2mm]
\textbf{
Ethan Caballero$^{2, 3}$,
Linbo Wang$^{4}$,
Ioannis Mitliagkas$^{2, 3}$,
Daniel M. Roy$^{4, 5}$
}
\\[3mm]
$^1$Element AI,
$^2$Mila,
$^3$Universit\'{e} de Montr\'{e}al,
$^4$University of Toronto,
$^5$Vector Institute\\
}

\maketitle

\begin{abstract}
One of the principal scientific challenges in deep learning is explaining generalization, i.e., why the particular way the community now trains networks to achieve small training error also leads to small error on held-out data from the same population. It is widely appreciated that some worst-case theories---such as those based on the VC dimension of the class of predictors induced by modern neural network architectures---are unable to explain empirical performance. A large volume of work aims to close this gap, primarily by developing bounds on generalization error, optimization error, and excess risk. When evaluated empirically, however, most of these bounds are numerically vacuous. Focusing on generalization bounds, this work addresses the question of how to evaluate such bounds empirically.
\citet{FGM} recently described a large-scale empirical study aimed at uncovering potential causal relationships between bounds/measures and generalization.
Building on their study, we highlight where 
their proposed methods can obscure 
failures and successes of generalization measures in explaining generalization.
We argue that generalization measures should instead be evaluated within the framework of distributional robustness.

\end{abstract}

\newcommand{\Vars}{\mathcal V}
\newcommand{\VarsProj}{\mathcal V^*}
\newcommand{\vv}{V}
\newcommand{\domain}[1]{D_{#1}}
\renewcommand{\env}{P}
\newcommand{\system}{\mathcal Q}
\newcommand{\cF}{\mathcal F}
\newcommand{\proj}[2]{#2|_{#1}}
\renewcommand{\Risk}[2]{R_{#1}(#2)}

\newcommand{\covar}[1][\env]{\bm X}
\newcommand{\resp}[1][\env]{\bm Y}
\newcommand{\icovar}[1][\env]{X}
\newcommand{\iresp}[1][\env]{Y}
\newcommand{\mScovar}[1][\env]{\Reals^{\pp \times \nn[#1]}}
\newcommand{\mSresp}[1][\env]{\Reals^{\times \nn[#1]}}
\newcommand{\Scovar}{\Reals^p}
\newcommand{\Sresp}{\Reals}
\newcommand{\ObsEnvs}{\mathcal E}
\newcommand{\UnobsEnvs}{\mathcal F}
\newcommand{\nn}[1][\env]{n_{#1}}
\newcommand{\pp}{p}
\newcommand{\diste}[1][\env]{P^{#1}}
\renewcommand{\env}{e}
\newcommand{\Sp}{\Omega}
\newcommand{\genmeas}{generalization measure\xspace}
\newcommand{\CM}{C}

\newcommand{\varvalues}{\omega}

\newcommand{\hh}{h_S}

\section{Introduction}

Despite tremendous attention, 
a satisfying theory of generalization in deep learning remains elusive.
In light of so many claims about explaining generalization in deep learning,
this statement is somewhat controversial.
It also raises an important question:
\begin{quote}
\emph{What does it mean to explain generalization in deep learning?}
\end{quote}
In this work, we propose empirical methodology to aid in the search of a precise mathematical theory,
allowing us to leverage large-scale empirical studies of generalization, like those in recent work \citep{jiang2018predicting,FGM}.
Unlike earlier work, however, our proposal rests on the foundation of \emph{robust prediction}, 
in order to catch out, rather than average out, failures.

The dominant approach to studying generalization is 
the frequentist framework of statistical learning theory.
We focus our attention on the simplest setting within supervised classification, 
where the training data, $S$, are modeled as a sequence of $n$ random variables, 
drawn i.i.d.\ from a distribution $\Dist$ on labeled examples $(x,y)$.
In supervised classification, 
learning algorithms choose a classifier $\hh$, based on the training data $S$.
Ignoring important considerations such as fairness, robustness, etc., 
the key property of a classifier $h$ is its probability of error, or (classification) risk,
\[\label{riskdefn}
\Risk{\Dist}{h}  = \Pr_{(x,y) \sim \Dist} [ h(x) \neq y ]. 
\]
One of the key questions is why deep learning often produces classifiers with human-level risk in domains that stymied researchers for decades. 
In this work, we take an empirical perspective and judge theories of generalizations by the predictions they provide when tested. In the other direction, any systematic rule for predicting generalization---whether learned or invented---can be thought of as a theory that can be tested.

We consider families of environments, defined by data distributions, architectural choices, train set sizes, learning algorithms and their tuning parameters, etc. 
Given a particular family of environments,
a strong theory achieves a desired level of precision for its predictions,
while 
depending as little as possible on the particular details of the environments.
At one extreme,
explanations based on the VC dimension of the zero--one loss class of neural networks 
would pin the success of deep learning on empirical risk minimization.
In practice, these explanations are poor, not just because the ensuing bounds are numerically vacuous for the size of networks and datasets used in practice, 
but because they fail to correctly predict the effect of changes to network width, depth, etc.

At the other extreme, 
average risk on held-out data (i.e., a test-set bound) provides a sharp estimate of risk,
yet the computation to produce this estimate is inextricably tied to every detail of the learned weights and data distribution. Viewing predictors as theories, the test-set bound is essentially silent.
Any satisfactory theory of generalization in deep learning must therefore lie between these two extremes.
We must necessarily exploit properties of the data distribution and/or learning algorithm,
but we must also be willing to trade precision to decrease our dependence on irrelevant details.

What dependence on the data distribution and learning algorithm is necessary
to explain deep learning?
Even taking the data distribution into consideration,
the fact that stochastic gradient descent (SGD) and its cousins often perform empirical risk minimization cannot explain generalization \citep{Rethinking17}.
There is, however, a picture emerging of overparametrization and SGD
 conspiring to perform capacity control.  In some theoretical scenarios, this control can be expressed in terms of norms.
At the same time, great strides have been made 
towards identifying notions of capacity that can be shown to  formally control generalization error,
$
\Risk{\Dist}{h} - \EmpRisk{S}{h}, 
$ 
uniformly over $h$ belonging to specially defined classes.
(Here $\EmpRisk{S}{h}$ denotes the empirical risk, as estimated by the training data.)

Despite this progress, there is still a wide gulf between 
performance as measured empirically via held-out data and 
performance as predicted by existing theory. %
No expert would be surprised to discover that a published bound %
yields predictions for risk that are numerically vacuous when evaluated empirically.
A standard retort is that the constants in most bounds are suboptimal
or that the purpose of bounds is to identify qualitative phenomena or inspire the development of new algorithms.
Even ignoring the issue of numerically vacuous bounds, 
many bounds demonstrably fail to account for the right dependencies.
As a case in point, 
recent empirical work \citep{NK19c} identifies state-of-the-art bounds on generalization error 
that grow with training set size, while the generalization error actually shrinks.
Indeed, many bounds are distribution- or data-dependent and so the question of whether they
explain generalization in practice is an empirical question.

\subsection{Large-scale empirical studies}

Recent work proposes large-scale empirical investigations to study generalization \citep{FGM}. (See also \citep{jiang2018predicting}.)
While it is becoming more common for theoretical work to present empirical evaluations, 
among recent empirical studies \citep[etc.]{GR17,neyshabur2018theRole,liao2018surprising,fisher-rao,bartlett2017spectrally,arora2018Compress}, most are limited.
One motivation for large-scale empirical studies is to leverage massive computing resources in the pursuit of a scientific challenge that has largely been approached mathematically.
Another motivation is to go beyond simply measuring correlation
towards measuring aspects of causation. 
(Several authors of this work---Dziugaite, Neal,  and Roy---have each advocated for this publicly.)

Given how influential these proposals by \citet{FGM} may be,
we believe they deserve critical attention. (Indeed, recent preprints have already started to integrate their methodology.)
\citet{FGM} propose to use Kendall correlation coefficients and independence tests to evaluate 
a suite of so-called \emph{generalization measures}.
Many of these generalization measures are frequentist bounds, though with missing constants or lower-order terms. 
Others are only loosely inspired by bounds.

One of central proposals made by \citeauthor{FGM} is to {\em average} evaluation metrics (i.e., Kendall-$\tau$) over a range of experimental settings, which arise from targeted changes to hyperparameters.%
\footnote{\citeauthor{FGM} also propose a conditional-independence test, which we discuss in \cref{app:jiang-ic}.}
In contrast, we argue that averaging across experimental settings is \emph{not} an appropriate summarization of the strength of a \genmeas
as a theory of generalization.
In particular, a satisfying theory should admit a \genmeas that offers reasonable predictions of generalization across the entire range of experimental settings under study.
A theory---realized by a generalization measure---is as strong as its weakest part: \textit{a satisfying theory cannot simply predict well on average}.

The study of prediction across a range of environments is the subject of \emph{distributional robustness}~\citep{arjovsky2019invariant,BuhlmannNL}.
An extreme form of robustness is obtained when one seeks to predict well in all environments that may arise from all possible interventions to an experimental setting. This extreme form of robustness can be linked to a weak form of causality \citep{BuhlmannNL, meinshausen2018causality}.

Crucially, we do not aim for robustness over all possible environments. 
To achieve some level of generality, useful theories must necessarily have limited scope.
As we demonstrate in \cref{svmexample}, frequentist generalization bounds can exploit noncausal correlations
that can be seen to stand in for unknown properties of the data distribution because of properties of the learning algorithm.
Such bounds have an important role to play in our search for a theory of generalization in deep learning,
but we cannot expect them to explain generalization under interventions that upset these noncausal correlations.
Bounds that depend on properties of the data distribution 
have an important role to play, though one hindered by the statistical barriers of unknown distributions, accessible only through a limited pool of data. More general theories (that minimize this dependence) can pinpoint key data properties.

\paragraph{Contributions.} 

Theories of generalization yield predictions:
How should we evaluate these predictions empirically? 
In this work, we adopt the proposal of \citep{FGM} to exploit large-scale empirical studies, but critique
the use of averaging in the evaluation of predictions made by these theories.
Based on the specific scientific goals of understanding generalization, 
we propose that the framework of distributional robustness is more appropriate, and suggest how to use it to evaluate generalization measures. 

Besides theoretical contributions, we make empirical contributions:
We collect data from thousands of experiments on CIFAR-10 and SVHN, with various values for width, depth, learning rate, and training set size. We adopt the ranking task and sign-error loss introduced by \citep{FGM},
but use the collected data to perform a \emph{robust} ranking evaluation, across the 24 candidate generalization measures on over 1,600,000 pairs of experiments.

We find that \emph{no existing complexity measure has better robust sign-error than a coin flip}.
Even though some measures perform well on average, 
every single measure suffers from 100\% failure in predicting the sign change in generalization error under some intervention. This observation is not the end of the evaluation, but the beginning.

To better understand the measures, we evaluate them in families of environments defined by interventions to a single hyperparameter. We find:
(i) most, though not all, measures are good at robustly predicting changes due to training set size; 
(ii) robustly predicting changes due to width and depth is hard for all measures, though some PAC Bayes-based measures are robust across a large fraction of the environments tested;
(iii) norm-based measures outperform other measures at learning rate interventions.

By focusing on robust evaluation, we force ourselves to dig into the data to uncover the cause of failures---failures which might otherwise go undiscovered by looking at average performance. As such, robustness provides better guidance to the scientific challenge of explaining generalization.
The rest of this paper is organized as follows. 
In \cref{svmexample}, we present a concrete example of a frequentist analysis of a learning algorithm, 
which reiterates some of the high-level points above.
We then introduce distributional robustness in \cref{sec:predstab} and describe how the framework can be used to analyze large-scale empirical studies of generalization measures in \cref{sec:gen-via-robust}.
We detail our experimental setting in \cref{sec:methods} and summarize our experimental findings in \cref{sec:empstudy} before ending with a discussion.

\newcommand{\HS}{\mathcal H}
\newcommand{\LossClass}{\mathcal L}

\section{A Motivating Example: SVMs, Norm-based Capacity Control, and the Role of Causality}
\label{svmexample}
In this section, we study support vector machines (SVMs) to demonstrate some of the challenges in understanding and explaining generalization error. This section owes much to \citep{shalev2014understanding}.
The intuition extracted from this simple model motivates our methodological choices for the rest of this paper. 
In particular, we see that frequentist generalization bounds derived under one set of conditions may rely on quantities that do not have a direct causal relationship with the generalization error under other conditions. This highlights that frequentist bounds can be expected to have limits to the predictive powers under intervention, but also that asking for causal measures of generalization may rule out measures that nonetheless work well in a range of scenarios.

\newcommand{\hww}{\overline{\ww}}
\newcommand{\wwS}{\ww_S}
\newcommand{\hwwS}{\overline{\ww_S}}
\newcommand{\oww}{\ww_*}
\newcommand{\howw}{\overline{\oww}}
Consider linear prediction, based on an embedding of inputs into $\Reals^p$. 
As usual, we index the space of linear predictors by nonzero vectors $\ww \in \HS = \Reals^p$,
where the decision boundary associated to $\ww$ is the 
tangent hyperplane $\{ \xx \in \Reals^p : \ip{\ww}{\xx} = 0 \}$, passing through the origin.
Assuming labels take values in $\{{\pm 1}\}$,
the zero--one classification loss of the predictor $\ww$ on a labeled input $(\xx,y)$ is 
$\loss(\ww,(\xx,y)) = \frac 1 2 (1 + y \sign(\ip{\xx}{\ww}))$.
Note that the loss is invariant to the magnitude of the vector $\ww$,
and so the set of hyperplanes can be put into correspondence with the unit vectors $\hww \defas \ww/\norm{\ww}$.
We focus on the realizable setting, i.e., data are assumed to be labeled according to some hyperplane.
In this case, every finite data set admits a positive cone of empirical risk minimizers.

Let $S=\{(\xx_i,y_i)\}_{i \in [n]}$ be $n$ i.i.d.\ labeled data in $\Reals^p \times \{{\pm 1}\}$,
and let $\wwS$ be chosen according to the SVM rule: 
$\min_{\ww} \|\ww\|^2$, subject to the constraint that $y_i \ip{\xx_i}{\ww} \ge 1$ for all $i \in [n]$. 
The constraint demands that, for each data point, 
the functional margin, $y_i \ip{\xx_i}{\wwS}$, be greater than one.
Thus the hyperplane $\hwwS$ indeed separates the data and achieves zero empirical risk.
However, among the vectors that satisfy the margin constraint, 
$\wwS$ has the smallest L2 norm.
Geometrically, the hyperplane $\hwwS$ is that with the largest \emph{geometric} margin, $\min_i y_i \ip{\xx_i}{\hwwS}$. 

Why does the SVM classifier generalize?
The best explanation may depend on the situation. 
The VC dimension of the space of $p$-dimensional linear predictors is $p$, and so,
with high probability over the sample $S$,
uniformly over all separating hyperplanes $\ww$,
the difference between the empirical risk and risk is $\tildeO(p/n)$. 
If $n \gg p$, then this reason alone suffices to explain strong performance. The details of the SVM rule are irrelevant beyond it returning an empirical risk minimizer.

Suppose that we consider a family of embeddings of growing dimensionality 
and find that the SVM rule generalizes equally well across this family. 
The VC theory cannot explain this. 
A theory based on the maximum-margin property of the SVM rule may.
To that end, assume there exists a hyperplane $\oww$ such that
$y \ip{\oww}{\xx} \ge 1$ with probability one over the pair $(\xx,y)$.
To fix a scale, assume $\norm{\xx} \le \rho$ with probability one.
By exploiting strong convexity, and the fact that $\norm{\ww_S} \le \norm{\oww}$,
one can show 
that the risk of $\wwS$ is bounded by $\tildeO(\rho \norm{\oww} / n)$.
Note that this bound has \emph{no explicit dependence} on the dimension $p$. 
Instead, it depends on the quantity $\rho \norm{\oww}$, whose reciprocal has a geometric interpretation: 
the distance between the separating hyperplane $\howw$ and the nearest data point, normalized by the radius of the data.
Therefore, this analysis trades dependence on dimension for dependence on the 
data distribution's density near the decision boundary.
When $\rho \norm{\oww} \ll n$, 
SVM's inductive bias is sufficient to explain generalization, even if $p \gg n$.

In fact, %
we can always build a bound based on the norm of the learned weights:
with high probability, for \emph{every} ERM $\wwS$,
the risk is bounded by $\tildeO(\rho \norm{\wwS} / n)$.
One might prefer such a bound since $\norm{\oww}$ is often presumed unknown.
Even if this bound matches risk empirically, it has a strange property: 
The bound depends on the norm $\|\wwS\|$ even though risk is \emph{independent of norm}.
Thus, we cannot expect the bound to remain valid if we intervene on the norm after training, e.g., to test for a causal relationship between norms and risk. Norms are the effect of the data and SVM interacting.

This example highlights that there may be multiple overlapping explanations depending on the range of environments in which one wants to understand generalization.
We cannot, however, expect a theory to be robust to arbitrary interventions.
Identifying a theory with limitations may lead us to more general ones, once we understand those limitations.
All of this motivates a careful design of experimental methodology, in order to navigate these tradeoffs.
In particular, we demand that a theory is \emph{robustly predictive} of generalization over a carefully designed family of \emph{environments}.

\newcommand{\SA}{\mathcal A}
\section{Preliminaries on Robust Prediction}
\label{sec:predstab}

In this section, we introduce  
the framework of robust prediction, 
borrowing heavily from \citet{BuhlmannNL,PetersEtAl2016,RothEtAl}. 
In the next section, 
we cast the problem of studying generalization into this framework.

Consider samples collected in a family $\UnobsEnvs$ of different \defn{environments}.
In particular, 
let $(\Sp,\SA)$ denote a common (measurable) sample space
and, in each of these environments $\env \in \UnobsEnvs$,
assume the data we collect are drawn i.i.d.\ from a distribution $\diste$ on $\Sp$. 
We will think of environments as representing different experimental settings, interventions to these experiments, sub-populations, etc.
For example, each sample might be a covariate vector and binary response, i.e., $\Sp = \Scovar \times \{{0,1}\}$.
A well-studied setting is where the distributions $\diste$ all agree on the conditional mean of the response given the covariates (i.e., the regression function), but disagree on the distribution of the covariates.

Prediction is formalized by a \defn{loss function}.
In particular, a loss function for a set $\Phi$ of predictors is a map $\loss : \Phi \times \Sp \to \Reals$. 
The \defn{error} or \defn{risk} (of a predictor $\phi \in \Phi$ in an environment $e \in \UnobsEnvs$)
is then the expected loss, $\EE_{\varvalues \sim \diste}[\loss(\phi,\varvalues)]$.
If we focus on one environment $\env \in \UnobsEnvs$, it is natural to seek a predictor $\phi \in \Phi$ with small risk \emph{for that individual environment}.
However, if we care about an entire family $\UnobsEnvs$ of environments,
we may seek a predictor that works well simultaneously across $\UnobsEnvs$.
In the setting of \defn{distribution robustness},
the performance of a predictor relative to a family $\UnobsEnvs$ of environments
is measured by the \defn{robust error (or risk)} 
\[\label{eq:robustrisk}\textstyle
\sup_{\env \in \UnobsEnvs} \EE_{\varvalues \sim \diste} [ \loss(\phi,\varvalues)].
\]
The goal of robust prediction is to identify a predictor with small robust risk.
\paragraph{Connection to causality.}
If taken to an extreme, then robust prediction is closely related to learning causality. Specifically, suppose that $(\icovar,\iresp)$ is induced by a common causal model $\iresp:=f(\icovar).$ If $\UnobsEnvs$ represents all possible interventions on subsets of $\icovar$, then the causal predictor $f(\icovar)$ also minimizes the robust risk. See \citep{rojas2018invariant,BuhlmannNL} for more details.

\section{Studying Generalization via Distributional Robustness}
\label{sec:gen-via-robust}
\newcommand{\singlenet}{single-network\xspace}
\newcommand{\Singlenet}{Single-network\xspace}
\newcommand{\couplednet}{coupled-network\xspace}
\newcommand{\Couplednet}{Coupled-network\xspace}

We are interested in understanding the effects of changes to a complex machine learning experiment, with a focus on effects on generalization.
In this section, we cast this problem into the framework of distributional robustness.
In order to study generalization,
we view theories of generalization as yielding predictors for generalization under a range of experimental settings.
We use the term \defn{generalization measure} to refer to such predictors.

\subsection{Experimental Records and Settings} 

In the notation of \cref{sec:predstab}, points $\varvalues \in \Sp$ represent possible samples. 
In our setting, each sample represents a complete record of a machine learning experiment.
An environment $\env$ specifies a distribution $\diste$ on the space $\Sp$ of complete records.

In the setting of supervised deep learning, a complete record of an experiment
would specify hyperparameters, random seeds, optimizers, training (and held out) data, etc.
Ignoring concerns of practicality, we assume the complete record also registers every conceivable derived quantity, not only including the learned weights, but also the weights along the entire trajectory, training errors, gradients, etc. 
Formally, we represent these quantities as random variables defined on the probability spaces $(\Sp,\SA,\diste)$, $\env \in \UnobsEnvs$. Among these random variables, there is the empirical risk $\hat R$ and risk $R$ of the learned classifier,
and their difference, $G$, the generalization error/gap.

Each distribution $\diste$ encodes the relationships between the random variables.
Some of these relationships are common to all the environments. 
E.g., the generalization error $G$ always satisfies $G = R - \hat R$, 
and the empirical risk $\hat R$ is always the fraction of incorrectly labeled examples in the training data.
Some relationships may change across environments.
E.g., in a family $\UnobsEnvs$ designed to  study SGD, 
changes to, e.g., the learning rate, affect the distribution of the trajectory of the weights.

In machine learning, environments arise naturally from learning algorithms applied to benchmarks under standard hyperparameter settings. 
In order to evaluate theories that explain the effect of, e.g., hyperparameter changes,
we also consider environments arising from perturbations/interventions to standard settings. 
E.g., we may modify
the hyperparameters or data, or intervene on the trajectory of weights in some way.
Every perturbation $\env$ is captured by a different distribution $\diste$. 
With respect to a family of environments $\UnobsEnvs$,
a generalization measure is preferred to another if it has smaller robust error (\ref{eq:robustrisk}). 
In \cref{sec:empstudy,sec:methods}, we restrict our attention to $\UnobsEnvs$ induced by varying 
hyperparameters, data distributions, training datasets, and dataset sizes.
In this work, we do not intervene on the dynamics of SGD.
However, intervening on the trajectory induced by SGD might be an interesting future direction that could allow one to tease apart the role of implicit regularization.

\subsection{Prediction tasks}
\label{sec:predtask}
The predictions associated with a theory of generalization are formalized in terms of a map $\GM : \Sp \to \Reals$,
which we call a \defn{generalization measure}.
We will study ad hoc generalization measures as well as ones derived from frequentist bounds.
In both cases, we are interested in the ability of these measures to predict changes in the generalization.

One important aspect of a generalization measure is the set of (random) variables (i.e., covariates) it depends on. 
Indeed, there is an important difference between the task of predicting generalization using only the architecture and number of data and using also, e.g., the learned weights.
Formally, let $V_1,\dots,V_k$ be a finite collection of random variables.
A generalization measure $\GM$ is \defn{$\sigma(V_1,\dots,V_k)$-measurable} if there exists a map $g$ such that
$\GM(\varvalues) = g(V_1(\omega),\dots,V_k(\omega))$ for all $\varvalues \in \Sp$.
We may prefer one generalization measure to another on the basis of the covariates it uses. As a simple example, if a generalization measure offers comparable precision to another measure, but is measurable with respect to a strict subset of variables, then this increased generality may be preferred.

\paragraph{Goals of the prediction.} 

We are broadly interested in two types of prediction tasks, distinguished by whether we train one or two networks.

In \emph{\couplednet} experiments, we train two networks, 
such that they share all hyperparameters except one.
We are interested in trying to predict which network has smaller generalization error.

Some of the generalization measures we consider are based on generalization bounds from the literature.
Given that generalization bounds are often numerically vacuous, it would not be informative to evaluate their predictions directly at this stage.
It is, however, reasonable to evaluate whether they capture the right dependencies.
Indeed, one desirable property of evaluating generalization measures by the rankings they induce in \couplednet experiments is that the rankings are invariant to monotonically increasing transformations of the measure.

In \emph{\singlenet} experiments, 
we try to predict the numerical value of the generalization error for that network based on a linear or affine function of a generalization measure.
Generalization measures that perform well in such a task would serve as accurate predictors of generalization, and could be used for, e.g., model selection. However, such measures would not necessarily serve to be useful in generalization bounds. We describe the experimental details and results of \emph{\singlenet} experiments in \cref{sec:regression} due to space limitations.

\section{Experimental methodology}
\label{sec:methods}

In \couplednet experiments, we evaluate the \emph{ranking} that the generalization measure induces on training networks. 
The approach we describe here is a robust analogue of the Kendall-$\tau$-based approach advocated 
 by \citet{FGM}.\footnote{See Appendix~\ref{app:jiang-ic} for a comparison to \citet{FGM}'s conditional-independence-based approach.} 
 This change is deceptively minor.
We highlight the very different conclusions drawn using our methodology in \cref{sec:empstudy}.

\paragraph{Evaluation criterion.} In more detail,
recall that a \couplednet environment $\env$ determines a distribution $\diste$ on pairs $(\varvalues,\varvalues')$ of {\em variable assignments}, each representing a full record of an experiment.
We evaluate a \genmeas, $\CM$, and the realized generalization error, $G$, on both assignments, $\varvalues$ and $\varvalues'$.
We use the ranking of $C$ values to predict the ranking of $G$ values.
Then, the \defn{sign-error of a \genmeas $\CM$} for this task\footnote{In order to match \citep{FGM}, in all of our experiments, we train to a fixed level of cross entropy loss that also yields zero training error. Since $G = R - \hat{R}$, and $\hat{R} =0$, a prediction task that depends on changes in generalization error $G$ is equivalent to one that depends on changes in risk $R$.} 
is given by
\[\label{eqn:sign-error}\textstyle
\SignE{\diste}{\CM} =  \frac 1 2 \EE_{(\varvalues,\varvalues') \sim \diste} 
     \big[1 - \sign \big(G(\varvalues')-G(\varvalues) \big)\cdot\sign\big(\CM(\varvalues')-\CM(\varvalues) \big) \big].
\]
Given a family $\cF$ of \couplednet environments, 
the \defn{robust sign-error of a \genmeas $\CM$} is $\sup_{\env \in \cF} \SignE{\diste}{\CM}$.
The $\Psi$ summary proposed by \citet{FGM} is analogous to the average sign-error, $|\cF|^{
-1} \sum_{\env \in \cF} \SignE{\diste}{\CM}$.\footnote{We apply a $\frac{1 - \Psi}{2}$ transformation to obtain values in $\left[0, 1\right]$, where $1$ is achieved if $\Psi = -1$ and $0$ if $\Psi = 1$.}

In our experiments, we use a modification of the loss in \cref{eqn:sign-error} 
in order to account for Monte Carlo variance in empirical averages. 
We use a weighted empirical average, where the weight for a sample $(\varvalues,\varvalues')$ is calculated based on the difference in generalization error $\abs{G(\varvalues)-G(\varvalues')}$.
We discard samples for which the difference in generalization error is below the Monte Carlo noise level.
In effect, we control the precision to which we want our generalization measure to predict changes: when the difference is insignificant, we do not predict the sign.
See \cref{app:montecarlo} for the details on how we use the Monte Carlo variance to choose what environments are being considered.
Other details of data collection are described in \cref{sec:expdetails}.

\paragraph{Environments.} In our experiments, variable assignments ($\varvalues$) are pairs $(H,\sigma)$ of hyperparameter settings and random seeds, respectively. The hyperparameters are: learning rate, neural network width and depth; dataset (CIFAR-10 or SVHN),
and training set size. (See \cref{sec:expdetails} for ranges.)

Each environment $\env$ is a pair $(H,H')$ of hyperparameter settings that differ in the setting of \emph{one} hyperparameter (e.g., depth changes from $2 \rightarrow 3$ between $H$ and $H'$ and the remaining hyperparameters are identical).
The distribution $\diste$ for a pair $\env = (H,H')$ is the distribution of $(\omega, \omega')=((H,\sigma),(H',\sigma'))$, where the random seeds $\sigma, \sigma'$ are chosen uniformly at random.
That is, the expectation in \cref{eqn:sign-error} is taken only over a random seed.

\section{Empirical Findings}
\label{sec:empstudy}

\begin{figure}
    \centering
    \includegraphics[width=\textwidth]{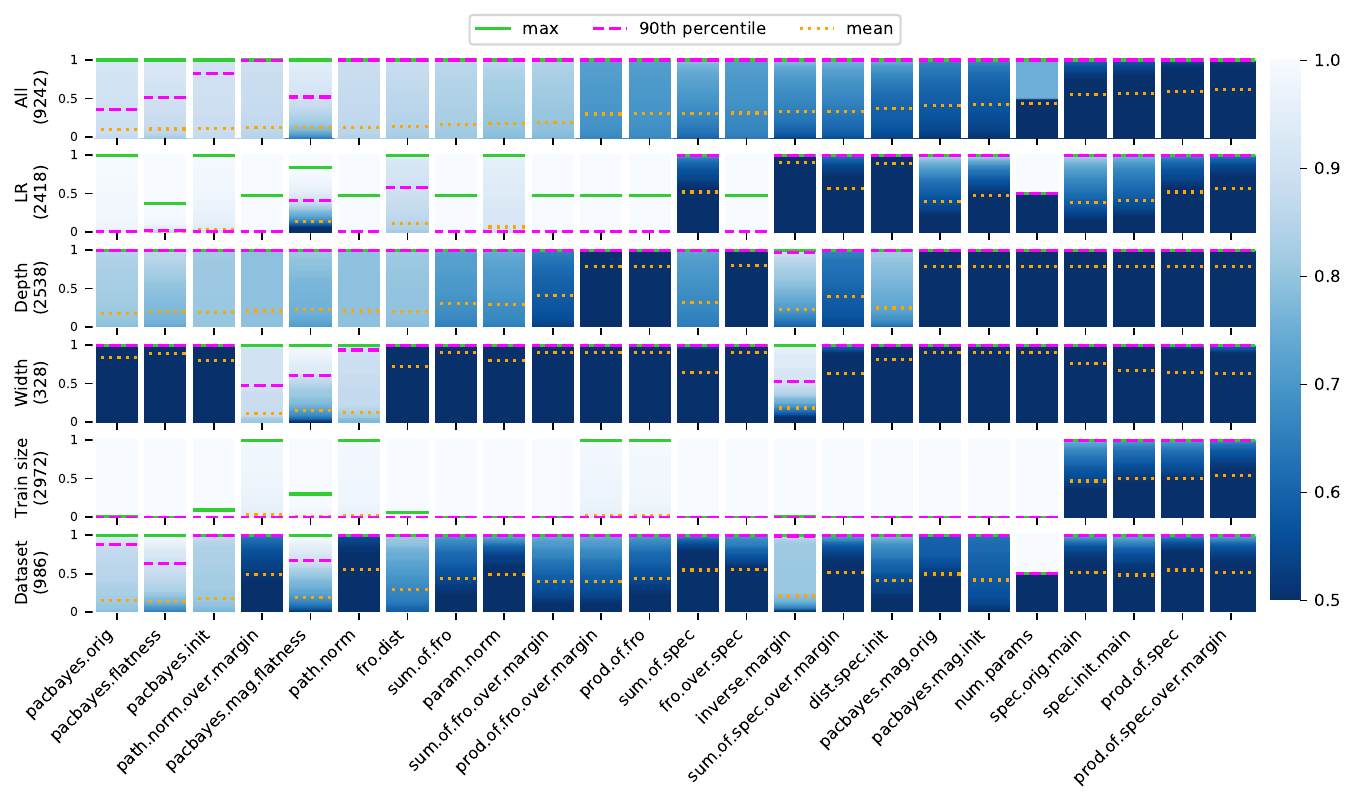}
    \caption{Cumulative distribution of the sign-error across subsets of environments for each generalization measure.
    The measures are ordered based on the mean across `All' environments.
    A completely \emph{white} bar indicates that the measure is perfectly robust, whereas a \emph{dark blue} bar indicates that it completely fails to be robust.}
    \label{fig:ranking_cifar10_svhn_cdf_per_param}
\end{figure}

In \cref{fig:ranking_cifar10_svhn_cdf_per_param}, we present a visualization of 1,600,000 ranking evaluations
on 24 generalization measures derived from those used in \citep{FGM}. A full description of these measures can be found in \cref{app:exp-measures}.
Motivated by the discussion in the introduction, we seek strong predictive theories: 
generalization measures that increase monotonically with generalization error and for which this association holds across a range of environments.
Such a measure would achieve zero robust sign-error (\cref{eqn:sign-error}).

As described in \cref{sec:methods}, each environment contains a pair of experiments that share all hyperparameters but one
(learning rate, depth, width, train set size, dataset).
In each environment, we calculate the weighted empirical average version of the sign-error 
over 100 samples from $\diste$ (10 networks runs with different seeds per $\varvalues$.
Note that we discard environments where too many samples have differences in generalization error below the Monte Carlo noise 
level (see \cref{sec:app-montecarlo-emp} for details). This is in contrast with the protocol proposed by \citet{FGM} where such noise is not filtered and
can significantly undermine the estimation of sign-error (see \cref{sec:app-montecarlo-ablation}).

In the remainder of this section we interpret the results of \cref{fig:ranking_cifar10_svhn_cdf_per_param},
highlight some significant shortcomings of the generalization measures, and point out cases where these shortcomings would have been obscured by 
non-robust, average-based summary statistics like those used by \citet{FGM}.

\paragraph{How to read \cref{fig:ranking_cifar10_svhn_cdf_per_param}.}
This figure presents the empirical cumulative distribution function (CDF) of the sign-error across all environments and 
\genmeas{}s. \textbf{Every row} shows the CDF over a subset of environments (e.g., those 
where only depth is varied). The `All' row  shows the same but over all environments. The number of environments in 
each subset is given on the left of each row. \textbf{Each bar} in the figure is the empirical CDF of all sign-errors in the set 
of environments. A bar's y-axis corresponds to the range of possible sign-errors and the internal coloring depicts the distribution
(starting at the median value for improved readability).
We annotate the bars with the max (i.e., robust sign-error; \textcolor{limegreen}{\textbf{green}}), 
the 90th percentile (\textcolor{magenta}{\textbf{magenta}}), and the mean (\textcolor{orange}{\textbf{orange}}).
The latter statistics do not measure robustness over all environments. 
However, a low 90th percentile value means the measure would have had 
low empirical robust sign-error restricting to some $90\%$ of the environments tested.
If the max is at 1.0, then there exists at least one environment where the measure fails to predict the sign of the change in 
generalization on all random seeds. If the max is below 0.5, then the measure is more likely than not to predict the correct 
sign on {\em all environments} in the set. 
\emph{Identifying subfamilies in which a measure is robust is one of our primary objectives.}

\paragraph{No measure is robust.}
As illustrated in the `All' row, for every one of the 24 measures,
there is at least one environment in which the measure \emph{always incorrectly predicts} the direction of change in generalization.
Nonetheless, some measures have low robust error over large fractions of environments,
as reflected by the 90th percentiles of the sign-error distributions.
Notice how the average-based summaries proposed by \citet{FGM} do not reflect robustness, 
which implies their inability to detect the causal associations that they seek.
Given these poor results, we must dig deeper 
to understand the merits and shortcomings of these generalization measures.
Therefore, we study their performance in natural subfamilies $\ObsEnvs \subseteq \UnobsEnvs$ of environments.
Our analyses of the `Train Size', `Depth', and `Width' rows below are examples of this. 
While no measure is robust across the CIFAR-10 and SVHN datasets considered here, we find measures 
that are quite robust over a $90\%$ fraction of environments
 for SVHN only (see \cref{app:couplednet-add-results-svhn}).

\paragraph{Robustness to train set size changes is not a given.}
In the `Train size' row, most measures correctly predict the effect of changing the train set size. 
(In general, generalization error decreases with train set size.)
It may seem a foregone conclusion that a bound of the form $\tildeO(\sqrt {c/n})$ would behave properly, but, 
for most of these measures, the complexity term $c$ is a random variable that can grow with more training data.
In fact, while many measures do achieve a low robust sign-error, 
some measures fail to be robust.
In particular, some bounds based on Frobenius norms (e.g., \texttt{prod.of.fro}; \cref{app:exp-measures-frobenius} and \citep{Ney1503})
increased with train set size in some cases.
Such corner cases arose mostly for shallow models (e.g., depth 2) with limited width (e.g., width 8) and were automatically identified by our proposed method. 
Note that the same finding was recently uncovered in a bespoke analysis \citep{NK19c},
and we may have missed this looking only at average sign-errors, which are usually low.

\begin{wrapfigure}{R}{0.35\textwidth}
    \centering
        \includegraphics[width=0.95\linewidth, trim=0cm .1cm 0cm .2cm, clip]{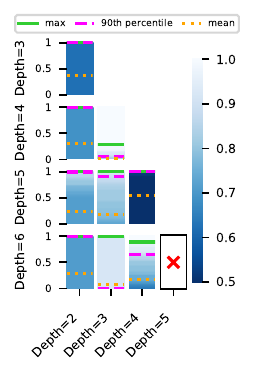}
    \caption{CDFs of sign-errors separated by environments where depth varies between two values for 
    \texttt{pacbayes.mag.flatness}. The red \textcolor{red}{X} indicates that no environments remained after accounting for Monte Carlo noise.}
    \label{fig:trianglecdf}
\end{wrapfigure}
\paragraph{Robustness to depth.} 
In the  `Depth' row, we depict robust sign-error for interventions to the depth.
Again, robust sign-error is maxed out for every measure. 
Digging deeper, these failures are not isolated: many measures actually fail in most environments.
However, there are exceptions: a few measures based on PAC-Bayes analyses show better performance in some environments.
In \cref{fig:trianglecdf}, we dig into the performance of \texttt{pacbayes.mag.flatness} (\cref{app:exp-measures-flatness}) by looking at the subset of environments where it performs well (e.g., varying depth 3 $\to$ 4),
fails but shows signs of robustness (e.g., 3 $\to$ 6), completely fails (e.g., 4 $\to$ 5), and those were
a conclusion cannot be reached (e.g., 5 $\to$ 6).
Looking into the data, we found that almost all environments where the measure fails are from the CIFAR-10 dataset,
where the smaller networks we test suffer from significant overfitting.
This illustrates how our proposed methodology can be used to zero-in on the limited scope where a measure is robust.

\paragraph{Robustness to width is surprisingly hard.}
In the `Width' row, all measures have robust sign-error close to 1.
Looking into the data, we discover that generalization error changes very little in response to interventions on width because the networks are all very overparametrized.
In fact, of the 4,000 available width environments, only 328 remain after accounting for Monte Carlo noise.

\paragraph{Comparison to \citet{FGM}}
Our contribution is primarily a methodological refinement to the proposals in \citep{FGM}. 
We describe how to discover failures
of generalization measures in specific environments by
looking at worst-case rather than average performance.
We note that there are several reasons that even our \emph{average-case} results are not directly comparable with those in \citep{FGM}.
First, their analysis considers the CIFAR-10 and SVHN datasets in isolation, whereas we combine models trained on both datasets.
Second, they do not account for Monte Carlo noise, which we found to significantly alter the distribution of sign-errors (see \cref{sec:app-montecarlo-ablation}).
This is important since we found that many environments had to be discarded due to high noise (e.g., only 8.2\% of the width environments remain after filtering out noise in our analysis).
Third, the hyperparameters and ranges that they consider are different from ours and, consequently, both studies look at different populations of models.
For example, the majority of  models in \citep{FGM} use dropout, whereas our models do not.
Such differences can alter how generalization measures and gaps vary in response to interventions on some hyperparameters and lead to diverging conclusions.
For instance, in our results, no measure has an average-case performance much better than a coin-flip in the `Depth' environments for CIFAR10, while \citet{FGM} 
find measures that perform well in this context.
Nevertheless, there are some general findings that persist across both studies;
for instance, we see the good average-case performance of the path-norm (\cref{app:exp-measures-path}) and PAC-Bayes-flatness-based (\cref{app:exp-measures-flatness}) measures in 
contrast to the poor performance of spectral measures (e.g., \texttt{prod.of.spec}; \cref{app:exp-measures-spectral}).
We also find more specific similarities, such as the poor average-case performance of most measures in `Width' environments for CIFAR-10 (\cref{app:couplednet-add-results-cifar10}), in contrast to
the better performance of \texttt{path.norm} (\cref{app:exp-measures-path}), \texttt{path.norm.over.margin} (\cref{app:exp-measures-path}), and \texttt{pacbayes.mag.flatness} (\cref{app:exp-measures-flatness}).

\section{Discussion}

The quest to understand and explain generalization is one of the key scientific challenges 
in deep learning.
Our work builds on recommendations in \citep{FGM} to use
large-scale empirical studies to evaluate generalization bounds. 
At the same time, 
we critique some aspects of these recommendations.
We feel that the proposed methodology in \citep{FGM} based on taking averages of sign-errors (or independence tests, which we have not pursued) 
can obscure failures. Indeed, for a long time, empirical work 
on generalization has not been systematic, and as a result, 
claims of progress outpace actual progress.

Based on an understanding of the desired properties of a theory of generalization,
we propose methodology that rests on the foundation of distributional robustness. 
Families of environments define the range of phenomena that we would like the theory 
to explain. A theory is then only as strong as its worst performance in this family.
In our empirical study, we demonstrated how a family can be broken down into subfamilies to help 
identify where failures occur.
While the present work focused on the analysis of existing measures of generalization, future work could build
on the robust regression methodology of Appendix~\ref{sec:regression} and attempt to formulate new robust measures
via gradient-based optimization.

The development of benchmarks and quantitative metrics has been a boon to machine learning. 
We believe that methodology based on robustness with carefully crafted interventions
will best serve our scientific goals.

\section*{Broader Impact}
 
Our work aims to sharpen our understanding of generalization 
by improving the way that we evaluate theories of generalization empirically.
The proposed methodology is expected to aid in the quest to understand generalization in deep neural networks.
Ultimately, this could lead to more accurate and reliable models and strengthen the impact of machine learning
in critical applications where accuracy must be predictable.
We believe that this work has no direct ethical implications.
However, as with all advances to machine learning, long-term societal impacts depend heavily on how machine learning is used.

\section*{Funding Sources}

LW was supported, in part, by an NSERC Discovery Grant.
DMR was supported, in part, by an NSERC Discovery Grant, Ontario Early Researcher Award, and a stipend provided by the Charles Simonyi Endowment. 
This research was carried out while GKD and DMR participated in the Special Year on Optimization, Statistics, and Theoretical Machine Learning at 
the Institute for Advanced Study.

\section*{Acknowledgements}

The authors would like to thank Grace Abuhamad, Alexandre Lacoste, Ga\"el Letarte, Ben London, Jeffrey Negrea, and Jean-Philippe Reid for feedback on drafts.

\printbibliography

\clearpage
\appendix
\addcontentsline{toc}{section}{Appendix} %
\part{Appendix} %
\parttoc %
\tableofcontents

\section{Importance sampling schemes to account for Monte Carlo noise}
\label{app:montecarlo}

 Generalization error $G(\varvalues)$ is estimated from a held-out test set, since we do not have access to the data distribution. 
 The size of the test set determines the precision at which true generalization error can be approximated.  
 Let $G, G'$ denote true generalization errors for $\varvalues,\varvalues'$, respectively, and similarly $\hat G, \hat{G'}$ estimates of generalization made on a test set of size $m$. 
 Let $\epsilon = \abs{\hat G-\hat G'}/2 $.
Then 
\begin{align}
\Pr ( \sign (G-G')  = \sign (\hat G- \hat{G'}) ) &\geq \Pr ( \abs{G- \hat G} \leq \epsilon) \cap (\abs{\hat{G'} - G'} \leq \epsilon) ) \\
&\geq ( 1 - \Pr (\abs{ G- \hat G} \geq \epsilon) ) (1 - \Pr(\abs{\hat{G'} - G'} \geq \epsilon) ).
\end{align} 
 We can now apply Hoeffding inequality to bound $\Pr ( \abs{G- \hat G} \geq \epsilon) \leq 2e^{-2m \epsilon^2}$.
 Let $\lossweight(\epsilon,m)  = (1-2e^{-2m\epsilon^2})^2$. 
If $\hat G$ and $\hat{G'}$ are computed using $m$ samples, then accepting samples only when $\lossweight(\epsilon,m)$ > p would mean that with probability at least $p$, $ \sign (G-G')  = \sign (\hat G- \hat{G'}) $. 

In our experiments, we only keep the samples with $\lossweight((\hat G(\varvalues')-\hat G(\varvalues))/2,m) >0.5$.
The expectation in the sign-error in \cref{eqn:sign-error}  is approximated by a weighted average of the loss for each sample, where the weight is equal to a rescaled version of
\[
\label{finalweight}
\losstransform (\varvalues, \varvalues') =  \max ( 0, \lossweight((\hat G(\varvalues')-\hat G(\varvalues))/2 ,m) -0.5).
\]

\subsection{Filtering environments}\label{sec:tau-filter-details}

The weighting scheme proposed in \cref{finalweight} allows to downweigh (and in some cases discard) pairs of experiments for 
which the generalization errors do not differ significantly. 
Consequently, some environments
may be left with very few samples. 
Let the $i^{\text{th}}$ sample have a weight $\kappa_i$.
To avoid calculating the expected sign-error on too few data points, we discard
environments where the \emph{effective sample size}, defined as
\[\label{effsamplesize}
\neff = \frac{  (\sum_{i} \kappa_i)^2}{\sum_i \kappa_i^2},
\]
is smaller than 12.
In our case, the weights are as defined in \cref{finalweight}. The choice of 12 samples means we estimate the expected loss to a precision of around standard deviation divided by 3 (\citep{mackay2003information}[Ch. 29, p. 380]).

\begin{figure}[h]
    \vspace{1cm}
    \centering
    \includegraphics[scale=0.7]{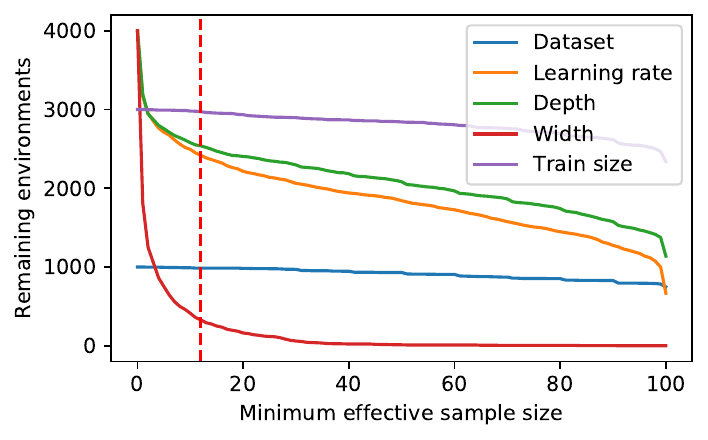}
    \caption{Number of environments remaining (per hyperparameter) at various effective sample size thresholds.
             The \textcolor{red}{red} vertical bar marks $\neff = 12$, which is the minimum effective sample size that we consider.
             Notice how most environments where width is varied have small effective sample sizes.}
    \label{fig:tau-n-envs}
\end{figure}

In \cref{fig:tau-n-envs}, we show the number of environments remaining at 
various $\neff$ cutoffs. Notice that very few environments are included for the width hyperparameter, even 
at $\neff \geq 12$. 
This is because the variations in generalization error due to width are often negligible in our data. 
Therefore, many of the environments where the width is varied are automatically discarded.

\subsection{Calculating the sign-error in empirical evaluation}\label{sec:app-montecarlo-emp}

\newcommand{\EmpSignE}[2]{\widehat{\mathrm{SE}}(#1,#2)}

Given the weighing and filtering schemes described above, we present the exact formulation of the sign-error used in the \couplednet experiments of \cref{sec:empstudy}.
For each environment $e$, 
we sample seeds $\{\sigma_i,\sigma_i'\}_{i \in \{1,..,10\} }$, 
yielding a total of 100 samples $\{\varvalues_i, \varvalues_j'\}_{(i,j) \in \{1,..,10\}^{\otimes 2} }$. For the sake of computation, we then use $\mathcal{E} = \{ (\varvalues_i, \varvalues'_i) \}_{(i,j) \in \{1,..,10\}^{\otimes 2}}$ as our samples from $\diste$.
We calculate the empirical sign-error as:
\[\label{eqn:sign-error-modified}\textstyle
\begin{split}
\EmpSignE{\mathcal{E}}{\CM} = 
    \frac 1 2 &\mathbbm{1} \Big( \neff  \geq 12\Big) \\
    & \sum_{(\varvalues,\varvalues') \in \mathcal{E}} 
     \bar\losstransform (\varvalues, \varvalues')
     \big[1 - \sign \big(\hat G(\varvalues')-\hat G(\varvalues) \big)\cdot\sign\big(\CM(\varvalues')-\CM(\varvalues) \big) \big],
\end{split}
\]
where the effective sample size $\neff$ is defined in \cref{effsamplesize}, 
and
$\bar\losstransform(\varvalues, \varvalues') = \dfrac{\losstransform (\varvalues, \varvalues')}{\sum_{(\varvalues_i,\varvalues'_i) \in \mathcal{E}} \losstransform (\varvalues_i, \varvalues_i')}$ are normalized weights.

\subsection{Ablation study: What is the effect of Monte Carlo noise?}\label{sec:app-montecarlo-ablation}

In this section, we investigate how filtering Monte Carlo noise using the aforementioned procedure affects the
estimated sign-errors. This is done by running an ablation study where the sign-errors are computed with and without 
Monte Carlo noise filtering and the resulting error distributions are compared. Specifically, for the \emph{with filtering} case,
pairs $(\varvalues, \varvalues')$ in each environment are weighted using $\losstransform(\varvalues, \varvalues')$ as 
described at \cref{finalweight}. Conversely, for the \emph{without filtering} case, pairs are weighted with 
$\losstransform(\varvalues, \varvalues') = 1$, allowing pairs with very small differences in generalization gap to be
included in the expectation. Intuitively, Monte Carlo noise should increase the sign-error in environments where it occurs,
since it can randomly push $\sign(\hat G(\varvalues) - \hat G(\varvalues'))$ to be $+1$ or $-1$, making it unpredictable.

\begin{figure}
    \centering
    \begin{subfigure}[b]{0.85\textwidth}
        \centering
        \includegraphics[width=\textwidth, trim=0 8 0 0, clip]{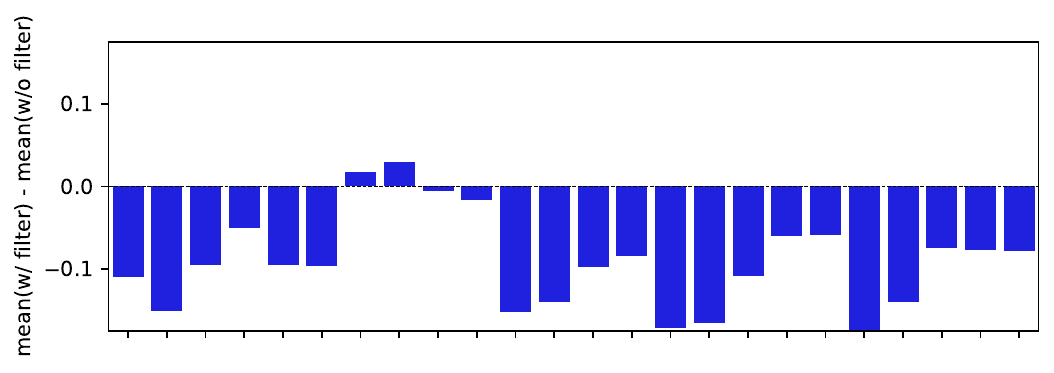}
        \caption{Mean}
        \label{fig:montecarlo-noise-ablation-mean}
    \end{subfigure}
    \begin{subfigure}[b]{0.85\textwidth}
        \centering
        \includegraphics[width=\textwidth, trim=0 8 0 0, clip]{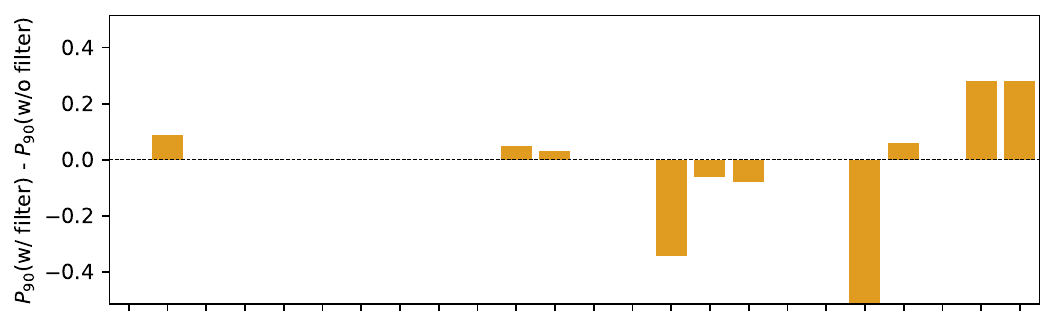}
        \caption{90$^\text{th}$ percentile}
        \label{fig:montecarlo-noise-ablation-90th}
    \end{subfigure}
    \begin{subfigure}[b]{0.85\textwidth}
        \centering
        \includegraphics[width=\textwidth, trim=0 8 0 0, clip]{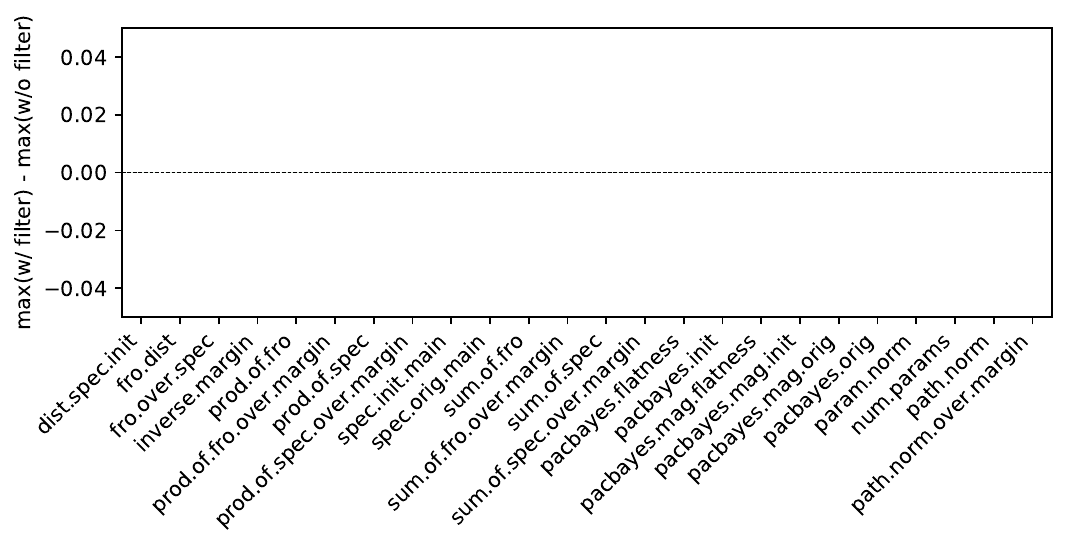}
        \caption{Maximum}
        \label{fig:montecarlo-noise-ablation-max}
    \end{subfigure}
       \caption{Effect of Monte Carlo noise filtering on various statistics of the sign-error distribution for each measure.}
       \label{fig:montecarlo-noise-ablation}
\end{figure}

The results reported in \cref{fig:montecarlo-noise-ablation} support this hypothesis: including noisy 
ranking pairs generally leads to larger sign-errors. 
Indeed, for many generalization measures, the mean sign-error decreases with noise filtering (\cref{fig:montecarlo-noise-ablation-mean}).
Moreover, extreme values such as the maximum do not seem to be affected by noise filtering (\cref{fig:montecarlo-noise-ablation-max}), which 
was expected since noise is likely not prevalent in each of the considered environments.
In light of these results, we conclude that the procedure that we propose to account for Monte Carlo noise
(\cref{app:montecarlo}) is beneficial and that it should be implemented in studies, such as ours and \citet{FGM},
that rely on ranking comparisons of generalization gaps.

\clearpage
\section{Evaluating robust prediction of the numerical value of generalization error}
\label{sec:regression}

In \singlenet experiments, we evaluate the ability of \defn{generalization measures} to predict the exact numerical 
value of the generalization error. 

\paragraph{Evaluation criterion.} We rely on a robust mean squared error (MSE) objective. 
For a transformation $f_{\theta}(\cdot)$ of a generalization measure $C$, the robust MSE is
\[\label{eq:affinerobustrisk}
\inf_{\theta} \, \sup_{\env \in \UnobsEnvs} \, \MSE{\diste}{f_{\theta} (\GM)}
\,\, \text{where $\MSE{\diste}{\CM'} = 
     \EE_{\varvalues \sim \diste} (\CM'(\varvalues) - \GenError(\varvalues))^2$.}
 \]
Choosing $f_a(x) = ax$ in \cref{eq:affinerobustrisk}, we recover \defn{robust risk of the linear oracle transformation 
of a generalization measure $\GM$}. Similarly, choosing $f_{(a,b)}(x) = ax+b$, we get \defn{robust risk of the affine 
oracle transformation of a generalization measure $\GM$}.

\paragraph{Environments.} In this setting, each environment $e \in \UnobsEnvs$ is defined by a single hyperparameter configuration $H$.
The data points in each environment are acquired by training a model with hyperparameters $H$ and varying the random seed.
For example, an environment could be composed of multiple experimental records where learning rate is $0.01$, the model 
depth is $2$, model width is $10$, the dataset is CIFAR-10, and the training set size is $50\,000$.
The hyperparameter values considered are those given at \cref{sec:expdetails} and we consider ten random seeds, 
resulting in 1000 environments with ten data points each.

\subsection{Experiments and results}\label{app:sec-reg-main-results}

\begin{figure}[h]
    \vspace{5mm}
    \centering
    \includegraphics[width=\textwidth]{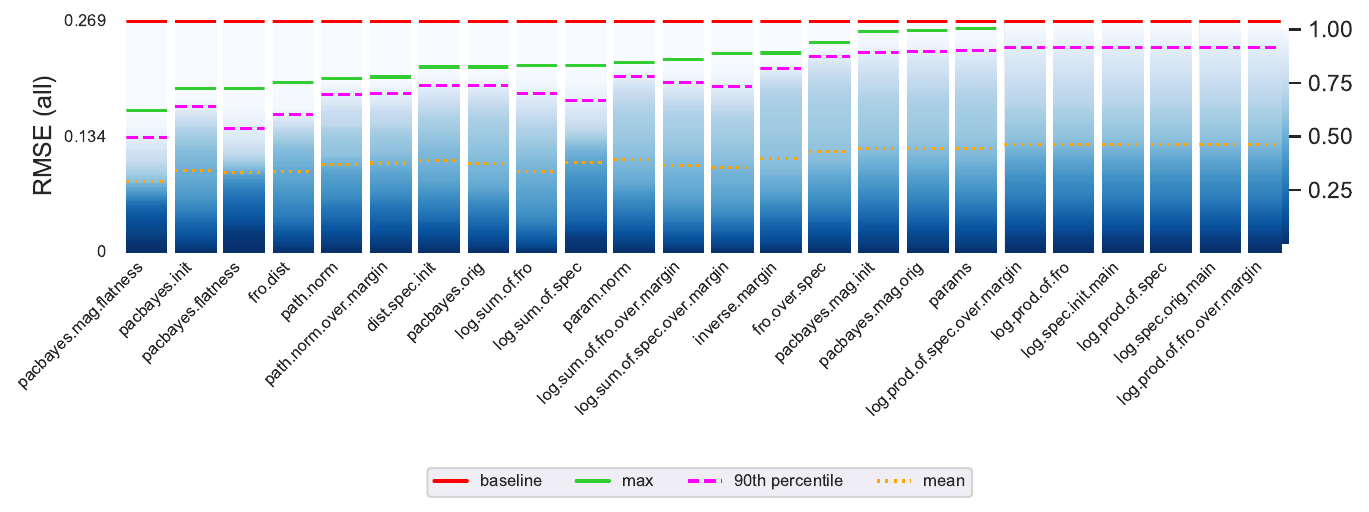}
    \caption{Cumulative distribution of robust root mean squared error (RMSE) of affine oracles trained on the $1000$ environments defined at \cref{app:sec-reg-main-results}.
    The maximum and average RMSE over all environments are shown in black and orange, respectively. A measure leading to
    perfect prediction of generalization error would have a white column and a black line at zero.
    The RMSE of a baseline oracle that only fits a bias parameter is shown in red.}
    \label{fig:regression-cdf-v1}
    \vspace{5mm}
\end{figure}

In our experiments, we fit an \emph{affine oracle} for each generalization measure by minimizing the robust mean squared error.
This evaluates the ability of each measure to predict generalization error in each environment, up to a common linear rescaling
and an additive constant.
Note that we constrain the linear coefficients to be non-negative (i.e., $a \geq 0$), since we expect these measures to upper bound generalization error.
We compare the performance of the affine oracles to that of a baseline that ignores the generalization measures and 
only fits a bias parameter (i.e., $a = 0$).
The results are reported in \cref{fig:regression-cdf-v1}.

Observe that many generalization measures achieve lower robust mean squared error than the baseline oracle.
This suggests that these measures do carry meaningful information about generalization error that transfers across all
environments.
In addition, notice that many measures that show good robust performance in this setting often perform well in the
\couplednet experiments (see \cref{sec:empstudy} and \cref{app:couplednet-add-results}).
For instance, the PAC-Bayes measures based on \texttt{flatness} show signs of robustness (although not perfect)
in both types of experiments.
Furthermore, measures based on \texttt{path.norm}, which show strong signs of robustness in the SVHN-only \couplednet
experiments (see \cref{app:couplednet-add-results-svhn}) are also among the best performers in this setting.
Finally, notice how the average mean squared error tends to be very similar across measures, while their robust mean
squared errors differ more.
This provides additional evidence that averaging can mask failures in robustness and that a worst-case analysis should
be preferred.

\subsection{Exploring weaker families of environments}

In this section, we consider families of environments $\UnobsEnvs$ which are intermediate between the robust regression described above and empirical risk minimization (where the average MSE is minimized).
Minimizing the robust risk should be an easier task in these environments, due to increasing levels of averaging which mask robustness failures.

\begin{figure}[h]
    \centering
    \includegraphics[width=\textwidth]{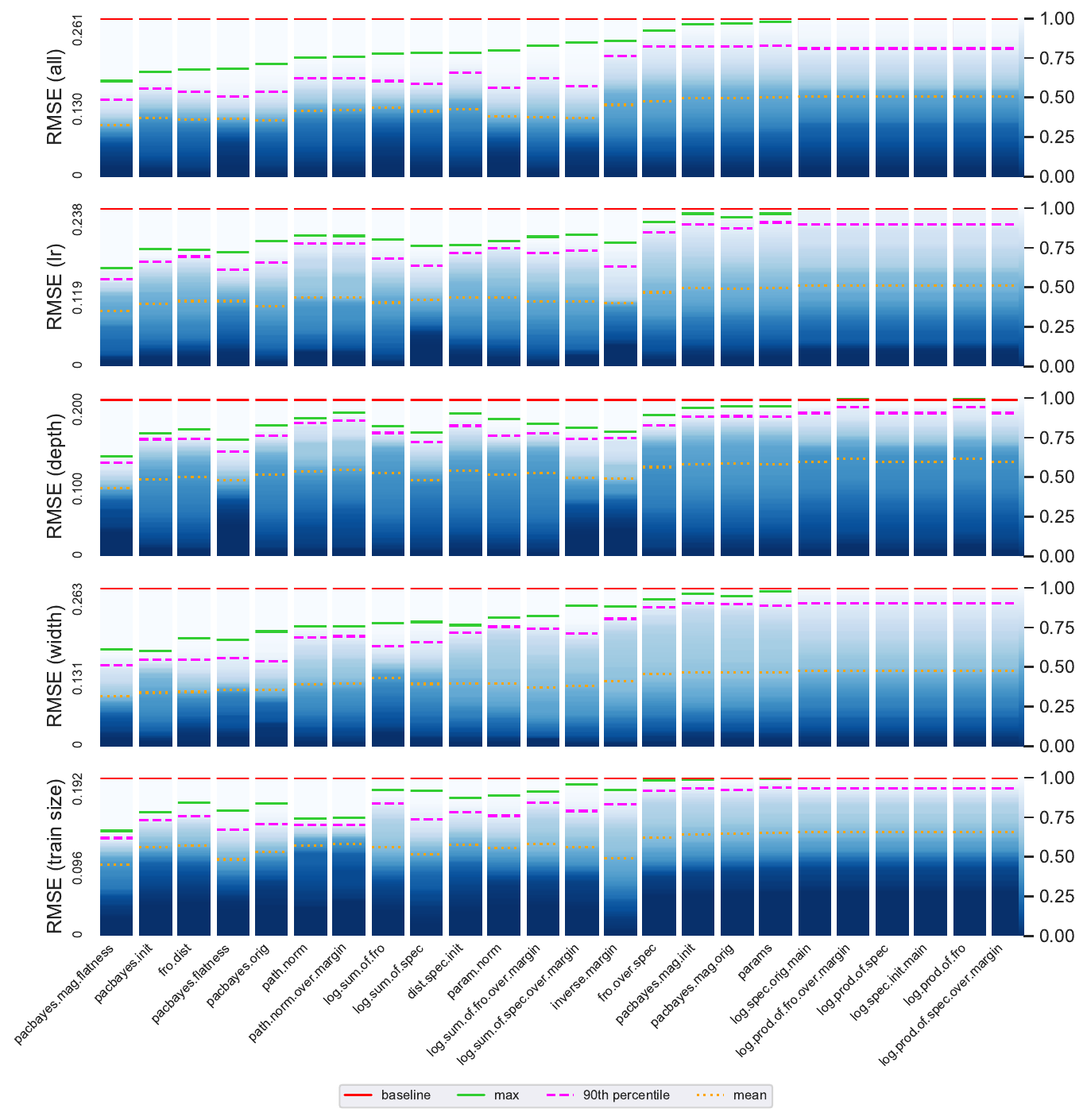}
    \caption{Cumulative distribution of robust root mean squared error (RMSE) of affine oracles trained on the $1350$ environments where a single 
    hyperparameter is varied and the others are fixed.
    Results are shown over all environments (top row) and separately for the subset of environments where each hyperparameter is varied (other rows).
    The maximum and average RMSE over all environments in each row are shown in black and orange, respectively. A measure leading to
    perfect prediction of generalization error would have a white column and a black line at zero.
    The RMSE of a baseline oracle that only fits a bias parameter is shown in red.}
    \label{fig:regression-cdf-v2}
\end{figure}

\paragraph{Varying a single hyperparameter.}
In this setting, each environment $e \in \UnobsEnvs$ is composed of runs where a single variable $H_i \in H$ is allowed to vary.
All other variables $\Vars \setminus H_i$ are fixed, though the random seed varies.
For example, an environment could be composed of all runs where the model depth is 2, model width is 10, the dataset is 
CIFAR-10, the training set size is $50\,000$, and the learning rate takes any of the $5$ values considered.
When considering the hyperparameter values described at \cref{sec:expdetails}, we obtain 1350 environments with between 20 to 50 points each (due to 10 repeats per hyperparameter setting).
The results for all experiments over this family of environments are reported in \cref{fig:regression-cdf-v2}.

As expected, the range of robust MSE values attained per measure is lower than in \cref{fig:regression-cdf-v1}, since the added averaging makes the task easier.
As in the previous section, we can see that the mean MSE over environments obscures a clear ordering over measures that is present when using the robust MSE.
Also, notice that the ordering induced by the robust MSEs is similar to that in \cref{fig:regression-cdf-v1}, with some PAC-Bayesian measures and \texttt{path.norm} measures performing best.

\begin{figure}[h]
    \centering
    \includegraphics[width=\textwidth]{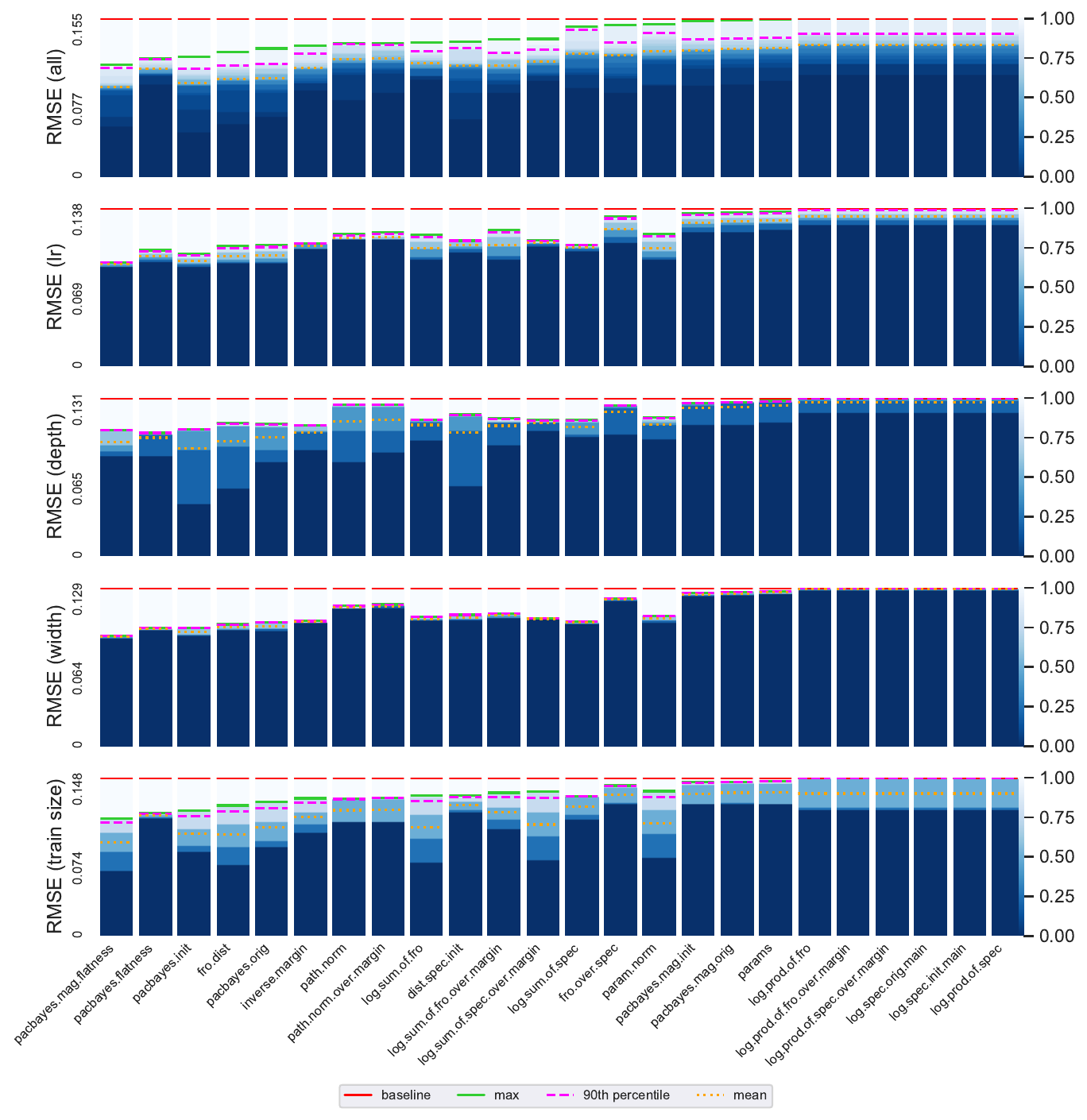}
    \caption{Cumulative distribution of robust root mean squared error (RMSE) of affine oracles trained on the $21$ environments where one hyperparameter
    is kept fixed and all the others are varied.
    Results are shown over all environments (top row) and separately for the subset of environments where each hyperparameter is kept fixed (other rows).
    The maximum and average RMSE over all environments in each row are shown in black and orange, respectively. A measure leading to
    perfect prediction of generalization error would have a white column and a black line at zero.
    The RMSE of a baseline oracle that only fits a bias parameter is shown in red.
    }
    \label{fig:regression-cdf-v3}
\end{figure}

\paragraph{Varying all but one hyperparameter.}
In this setting, each environment $e \in \UnobsEnvs$ is composed of runs where a single variable $H_i \in H$ is fixed.
All other variables $\Vars \setminus H_i$ and the random seed vary.
For example, an environment could be composed of all runs where the model depth is 2 and the other parameters take on every possible value.
When considering the hyperparameter values described at \cref{sec:expdetails}, we obtain 21 environments with 2000 to 5000 points each.
The results for all experiments over this family of environments are reported in \cref{fig:regression-cdf-v3}.

The gap between the mean and robust RMSEs is much narrower here, caused by averaging the MSE over much more points.
Nevertheless, we are still able to see a very similar ordering to that of \cref{fig:regression-cdf-v1} and \cref{fig:regression-cdf-v2} preserved here.

\clearpage
\section{Experimental details, code, and data}
\label{sec:expdetails}
Our experimental protocol is inspired by that used in the large-scale study of \citet{FGM}.
We proceed in two steps: 1) data collection, where many models are trained with various hyperparameter configurations, 2) data analysis (e.g. coupled-network experiments; see \cref{sec:methods}).
Our code relies on PyTorch \cite{PyTorch}, NumPy \cite{harris2020array}, Pandas \cite{mckinney-proc-scipy-2010}, Scikit-Learn \cite{scikit-learn} and Matplotlib \cite{Hunter:2007}.

\paragraph{Availability:} Our code is open-source and available online, along with the data used in the experiments: \href{https://github.com/nitarshan/robust-generalization-measures}{https://github.com/nitarshan/robust-generalization-measures}

\subsection{Hyperparameters}
\label{app:exp-hparams}
Each data point in our analysis is obtained by training a model with a given hyperparameter configuration. 
Between data points, we vary 5 hyperparameters that alter the model, the learning procedure, and the data distribution.
These are:

\begin{enumerate}
    \item Learning rate: $H_{lr} \in \{10^{-3}, 10^{-2.8}, 10^{-2.5}, 10^{-2.2}, 10^{-2}\}$
    \item Model depth: $H_{depth} \in \{2, 3, 4, 5, 6\}$ \\[1mm]
        \emph{This corresponds to the number of "blocks" used in our model architecture. Each block corresponds to 3 convolutional layers.}
    \item Model width: $H_{width} \in \{8\times 25, 10\times 25, 12\times 25, 14\times 25, 16\times 25\}$ \\[1mm]
        \emph{This corresponds to the filter width used in convolutional layers of our model architecture.}
    \item Dataset: $H_{dataset} \in \{\text{CIFAR-10}, \text{SVHN}\}$
    \item Training set size: $H_{size} \in \{ 50\,000, 25\,000, 12\,500, 6\,250 \}$
\end{enumerate}

\subsection{Models}
\label{app:exp-models}
We use a fully convolutional "Network-in-Network" architecture similar to that described in \cite{networkinnetwork} and used for the study in \cite{FGM}. A full specification of our model can be found in our codebase.

While the most successful model architectures of today employ residual connections between blocks of convolutional layers, we are unable to make use of those here due to the unclear applicability of many bounds to models using skip-connections.

\subsection{Datasets}
\label{app:exp-datasets}
We make use of two common vision datasets: CIFAR-10 \cite{CIFAR10} and SVHN \cite{SVHN}. 
Both datasets are composed of 32x32 RGB images with 10 classes of natural images, with CIFAR-10's classes corresponding to animals and vehicles, and SVHN's classes corresponding to digits cropped from Street View images. 
We make use of the full training (50k images) and testing (10k images) splits of CIFAR-10, and randomly sample (without replacement) a subset of the larger training and testing sets of SVHN to match the split sizes of CIFAR-10. 
We also sample without replacement when generating the smaller training sets of size 25k, 12.5k and, 6.25k, but always make use of the same testing sets of size 10k across all experiments.

We do not make use of data augmentation when passing these images into our models, following the observation in \cite{FGM} that doing so negatively affects the ability of these models to consistently reach low cross-entropies.

\subsection{Training procedure}
\label{app:exp-training-procedure}
We use SGD with a momentum parameter of $0.9$ for all experiments. We do not use learning rate decay or weight decay. 
As in \cite{FGM} we use a cross-entropy stopping criterion, which we set to $0.01$ for all experiments, and calculate over the entire training dataset.

\subsection{Data collection}
\label{app:exp-data-collection}

We run 10 repeats with different random seeds for each of the $1000$ possible hyperparameter combinations, providing a total of $10,000$ experimental runs. 
Of these, 300 runs failed to meet the cross-entropy criterion as well as an additional training accuracy criterion of being greater than $99\%$.
These data points were filtered out before any analysis.

As in \cite{FGM} we "fuse" the batch-norm layers of our model with their preceding convolutional kernels before calculating the value of the generalization measures.

\subsection{Measures}
\label{app:exp-measures}
We look at the following 24 generalization measures for convolutional networks, which are modifications of a subset of those studied in \cite{FGM}. 
A key difference is that, while those original measures do not account for dataset size, we correct for this through a normalization of the form $\sqrt{C/m}$ for all measures $C$, where $m$ is the size of the training dataset. 

While we provide the mathematical expressions for all our measures here, we direct the reader to \cite[Appendix D]{FGM} for more details. It is worth noting, however, that while many of these expressions are derived from generalization bounds, there is no requirement that generalization measures correspond to bounds. 
Even among the measures that correspond to bounds, direct comparison is not necessarily meaningful.
Some are generalization bounds for the networks learned by SGD. Some, in particular those derived from PAC-Bayes methods, are bounds on stochastic classifiers, e.g., obtained by randomizing the weights of a neural network in some way. Some of these quantities control the generalization error between the risk and empirical risk, while others relate to the difference between surrogate risks, e.g., based on margins. 

We calculate the spectral norm of convolutional layers using the exact FFT-based method of \cite{sedghi2018the}.
We had initially used the approximation proposed in \citet{yoshida2017spectral}, but our findings were not noticeably different between the approximation and exact methods.

For reasons of numerical stability, we apply log transformations to some of these measures. \textbf{As this is a monotonic transformation, it does not affect the ranking results covered in the main paper.}

\subsubsection{VC Measures}
\label{app:exp-measures-vc}
For a convolutional network of $d$ layers, with a $k_i \times k_i$ kernel and $c_i$ filters at depth $i$:
\begin{equation}
    C_{params} = \sqrt{\frac{\sum_i^d k_i^2 c_{i-1} (c_i + 1)}{m}} \label{eq:params}
\end{equation}

\subsubsection{Output Measures}
Let $\gamma$ be the 10th-percentile of margin values over the training dataset.
\label{app:exp-measures-output}
\begin{equation}
    C_{inverse.margin} = \sqrt{\gamma^2 m}^{-1} \label{eq:inverse_margin}
\end{equation}

\subsubsection{Spectral Measures}
\label{app:exp-measures-spectral}
Let $\mathbf{W}_i$ denote the $i$\textsuperscript{th} convolutional layer's weight tensor, and $\mathbf{W}_i^0$ it's initial value. Let $\| \mathbf{W}_i\|_2$ denote it's spectral norm.
Let $\| \mathbf{W}_i\|_F = \| \text{vec}(\mathbf{W}_i) \|$ denote it's Frobenius norm.
\begin{align}
    C_{log.spec.init.main} &= \log\sqrt{\frac{\prod_{i = 1}^d \| \mathbf{W}_i\|_2^2 \sum_{j = 1}^d \frac{\| \mathbf{W}_j - \mathbf{W}_j^0 \|_F^2}{\| \mathbf{W}_j\|_2^2}}{\gamma^2 m}} \label{eq:log_spec_orig_main} \\
    C_{log.spec.orig.main} &= \log\sqrt{\frac{\prod_{i = 1}^d \| \mathbf{W}_i\|_2^2 \sum_{j = 1}^d \frac{\| \mathbf{W}_j\|_F^2}{\| \mathbf{W}_j\|_2^2}}{\gamma^2 m}} \label{eq:log_spec_orig_main} \\
    C_{log.prod.of.spec.over.margin} &= \log\sqrt{\frac{\prod_{i = 1}^d \| \mathbf{W}_i\|_2^2}{\gamma^2 m}} \label{eq:log_prod_of_spec_over_margin} \\
    C_{log.prod.of.spec} &= \log\sqrt{\frac{\prod_{i = 1}^d \| \mathbf{W}_i\|_2^2}{m}} \label{eq:log_prod_of_spec} \\
    C_{fro.over.spec} &= \sqrt{\frac{\sum_{i = 1}^d \frac{\| \mathbf{W}_i\|_F^2}{\| \mathbf{W}_i\|_2^2}}{m}} \label{eq:fro_over_spec} \\
    C_{log.sum.of.spec.over.margin} &= \log\sqrt{\frac{d \left( \frac{\prod_{i = 1}^d \| \mathbf{W}_i\|_2^2}{\gamma^2} \right)^{1/d}}{m}} \label{eq:log_sum_of_spec_over_margin} \\
    C_{log.sum.of.spec} &= \log\sqrt{\frac{d \left( \prod_{i = 1}^d \| \mathbf{W}_i\|_2^2 \right)^{1/d}}{m}} \label{eq:log_sum_of_spec}
\end{align}

\subsubsection{Frobenius Measures}
\label{app:exp-measures-frobenius}

\begin{align}
    C_{log.prod.of.fro.over.margin} &= \log\sqrt{\frac{ \prod_{i = 1}^d \| \mathbf{W}_i \|_F^2}{\gamma^2 m}} \label{eq:log_prod_of_fro_over_margin} \\
    C_{log.prod.of.fro} &= \log\sqrt{\frac{ \prod_{i = 1}^d \| \mathbf{W}_i \|_F^2}{m}} \label{eq:log_prod_of_fro} \\
    C_{log.sum.of.fro.over.margin} &= \log\sqrt{\frac{d \left( \frac{1}{\gamma^2} \prod_{i = 1}^d \| \mathbf{W}_i \|_F^2 \right)^{1/d}}{m}} \label{eq:log_sum_of_fro_over_margin} \\
    C_{log.sum.of.fro} &= \log\sqrt{\frac{d \left( \prod_{i = 1}^d \| \mathbf{W}_i \|_F^2 \right)^{1/d}}{m}} \label{eq:log_sum_of_fro} \\
    C_{fro.dist} &= \sqrt{\frac{\sum_{i = 1}^d \| \mathbf{W}_i - \mathbf{W}_i^0 \|_F^2}{m}} \label{eq:fro_dist} \\
    C_{dist.spec.init} &= \sqrt{\frac{\sum_{i = 1}^d \| \mathbf{W}_i - \mathbf{W}_i^0 \|_2^2}{m}} \label{eq:dist_spec_init} \\
    C_{param.norm} &= \sqrt{\frac{\sum_{i = 1}^d \| \mathbf{W}_i \|_F^2}{m}} \label{eq:param_norm}
\end{align}

\subsubsection{Path Measures}
\label{app:exp-measures-path}
Define the parameter vector as $\mathbf{w} = \text{vec}(\mathbf{W}_1,\dots,\mathbf{W}_d)$. Below, $f_{\mathbf{w}^2}(\mathbf{1})[i]$ denotes the $i$\textsuperscript{th} logit output of a network using squared weights where the input is a vector of ones.
\begin{align}
    C_{path.norm.over.margin} &= \sqrt{\frac{\sum_i f_{\mathbf{w}^2}(\mathbf{1})[i]}{\gamma^2 m}} \label{eq:path_norm_over_margin} \\
    C_{path.norm} &= \sqrt{\frac{\sum_i f_{\mathbf{w}^2}(\mathbf{1})[i]}{m}} \label{eq:path_norm}
\end{align}

\subsubsection{Flatness Measures}
Let $\omega$ denote the number of weights. Let $\epsilon = 1 \times 10^{-3}$.
We use the search procedure for $\sigma$ described in \cite{FGM},
where it is chosen to be the largest number such that $\EE_{\mathbf{u}\sim \mathcal{N}(0,\sigma^2 I)} \left[ \hat{L} (f_{\mathbf{w} + \mathbf{u}}) \leq 0.1 \right]$.
Similarly, we choose the magnitude-aware $\sigma'$ to be the largest number such that $\EE_{\mathbf{u}} \left[ \hat{L} (f_{\mathbf{w} + \mathbf{u}}) \leq 0.1 \right]$, where $u_i \sim \mathcal{N}(0,\sigma'^2 |w_i|^2 + \epsilon^2)$.
\label{app:exp-measures-flatness}
\begin{align}
    C_{pacbayes.init} &= \sqrt{\frac{\frac{\|\mathbf{w} - \mathbf{w}^0\|_2^2}{4 \sigma^2} + \log \left( \frac{m}{\sigma} \right) + 10}{m}} \label{eq:pacbayes_init} \\
    C_{pacbayes.orig} &= \sqrt{\frac{\frac{\|\mathbf{w}\|_2^2}{4 \sigma^2} + \log \left( \frac{m}{\delta} \right) + 10}{m}} \label{eq:pacbayes_orig} \\
    C_{pacbayes.flatness} &= \sqrt{\frac{1}{\sigma^2 m}} \label{eq:pacbayes_flatness} \\
    C_{pacbayes.mag.init} &= \sqrt{\frac{\frac{1}{4} \sum_{i = 1}^\omega \log \left( \frac{\epsilon^2 + (\sigma'^2 + 1) \|\mathbf{w} - \mathbf{w}^0\|_2^2 / \omega}{\epsilon^2 + \sigma'^2 |w_i - w_i^0|^2} \right) + \log \left( \frac{m}{\delta} \right) + 10}{m}} \label{eq:pacbayes_mag_init} \\
    C_{pacbayes.mag.orig} &= \sqrt{\frac{\frac{1}{4} \sum_{i = 1}^\omega \log \left( \frac{\epsilon^2 + (\sigma'^2 + 1) \|\mathbf{w}\|_2^2 / \omega}{\epsilon^2 + \sigma'^2 |w_i - w_i^0|^2} \right) + \log \left( \frac{m}{\delta} \right) + 10}{m}} \label{eq:pacbayes_mag_orig} \\
    C_{pacbayes.mag.flatness} &= \sqrt{\frac{1}{\sigma'^2 m}} \label{eq:pacbayes_mag_flatness}
\end{align}

\clearpage
\section{\Couplednet: Additional experimental results}
\label{app:couplednet-add-results}

\subsection{Restricting the analysis to the SVHN dataset}
\label{app:couplednet-add-results-svhn}

\begin{figure}[h]
    \centering
    \vspace{1cm}
    \includegraphics[width=0.9\textwidth]{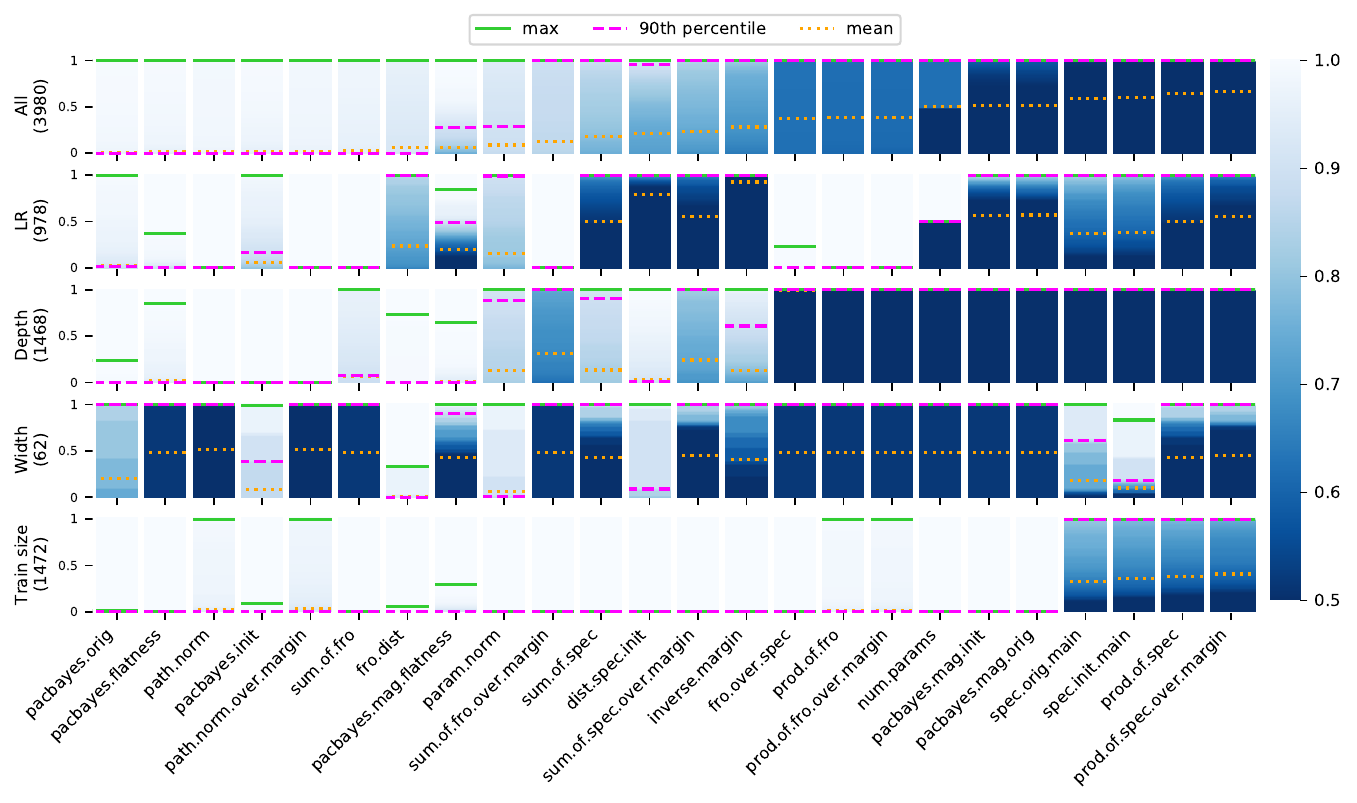}
    \caption{Cumulative distribution of the sign-error across subsets of environments for each generalization measure (in SVHN only).
    The measures are ordered based on the mean across `All' environments.
    A completely \emph{white} bar indicates that the measure is perfectly robust, whereas a \emph{dark blue} bar indicates that it completely fails to be robust.}
    \label{fig:ranking-cdf-svhn-c0}
\end{figure}

In \cref{sec:empstudy}, we observed that some measures failed to be robust across changes in depth for CIFAR-10 environments (e.g., \texttt{pacbayes.mag.flatness}). 
It is therefore reasonable to ask what would happen if we restricted the study to SVHN.
Thus, we replicate the experiments reported in \cref{fig:ranking_cifar10_svhn_cdf_per_param} of the main text, but leave
out all CIFAR-10 environments. The results are reported in \cref{fig:ranking-cdf-svhn-c0}. 

Notice how all measures still achieve a robust sign-error of $1.0$ overall, but that many measures now have a 90th percentile much closer to zero.
This indicates that the error distributions of some measures would have been judged to be robust on some large subfamily of environments.
Furthermore, observe how some measures now achieve perfect robustness on `Depth' environments (e.g., \texttt{path.norm}), while none had achieved a sign-error
lower than 1.0 in \cref{fig:ranking_cifar10_svhn_cdf_per_param}.

\begin{figure}
    \centering
    \begin{subfigure}[b]{0.45\textwidth}
        \centering
        \includegraphics[scale=1.3]{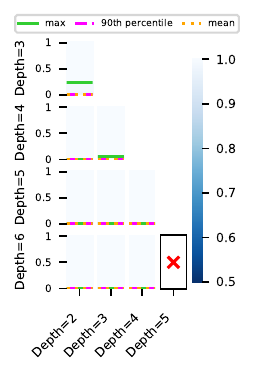}
        \caption{Interventions on depth}
        \label{fig:svhn-triangle-depth}
    \end{subfigure}
    \hfill
    \begin{subfigure}[b]{0.45\textwidth}
        \centering
        \includegraphics[scale=1.3]{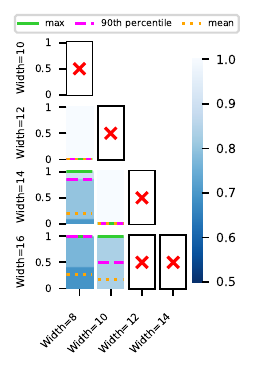}
        \caption{Interventions on width}
        \label{fig:svhn-triangle-width}
    \end{subfigure}
    \caption{CDFs of sign-errors separated by environments where depth and width vary between two values for 
    \texttt{pacbayes.orig} (on SVHN only). The red \textcolor{red}{X}s indicate that no environments remained after accounting for Monte Carlo noise.}
\end{figure}

Our methodology allows to dig deeper into these results.
For instance, we can try to understand where \texttt{pacbayes.orig} fails to be robust by looking at the CDF of sign-errors for every pair of values of 
depth (\cref{fig:svhn-triangle-depth}) and width (\cref{fig:svhn-triangle-width}).
We observe that most failures in `Depth' environments occur when varying depth from $2 \rightarrow 3$.
The sources of non-robustness in width are slightly harder to interpret, but we still observe that the
measure is significantly more robust for some pairs of width than others.

Digging even deeper, we can look at the distribution of hyperparameter values in the environments where this measure fails to be robust.
For `Depth' environments, we observe that there are 10 environments with a sign-error greater than $0.01$ and that these all correspond
to very wide networks ($\{14, 16\}$) with small learning rates ($\{0.001, 0.0016\}$) trained on a small dataset ($6250$ examples).
For `Width' environments, we observe $16$ environments with a sign-error greater than $0.01$ and that these all correspond to shallow
models of depth $2$, trained with large learning rates ($\{10^{-2.5}, 10^{-2.2}, 10^{-2}\}$) on datasets of less than $50,000$ examples.

This example clearly illustrates how our proposed methodology allows to zero-in on cases where generalization measures fail
to be robust. We expect that studying the shortcomings of measures at such a detailed level will aid in elaborating new, 
more robust, theories of generalization.

\clearpage

\subsection{Restricting the analysis to the CIFAR-10 dataset}
\label{app:couplednet-add-results-cifar10}

\begin{figure}[H]
    \centering
    \vspace{1cm}
    \includegraphics[width=0.9\textwidth]{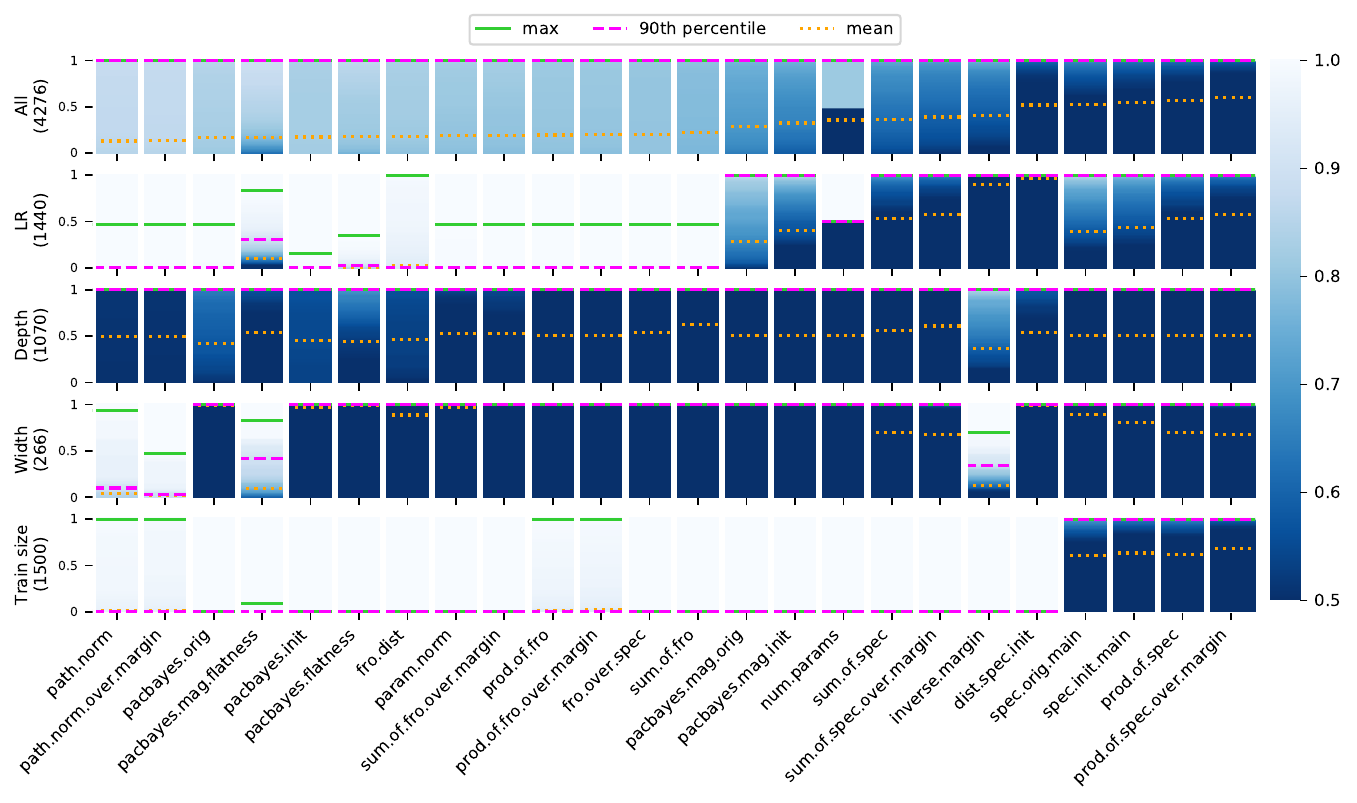}
    \caption{Cumulative distribution of the sign-error across subsets of environments for each generalization measure (in CIFAR-10 only).
    The measures are ordered based on the mean across `All' environments.
    A completely \emph{white} bar indicates that the measure is perfectly robust, whereas a \emph{dark blue} bar indicates that it completely fails to be robust.}
    \label{fig:ranking-cdf-cifar10}
\end{figure}
\clearpage

\subsection{Exploring a weaker family of environments}

We further evaluate the performance of generalization measures in a family of environments that is intermediate between the one considered in \cref{sec:methods} and by \citet{FGM}.
As described in \cref{sec:methods}, each environment contains pairs of hyperparameter settings where, within a pair, one hyperparameter is varied between two specific values (e.g., depth $2 \rightarrow 3$).
However, here, the value of the other hyperparameters is allowed to change between the pairs.
For example, assuming only two hyperparameters (width ($w$) and depth ($d$)), the pairs $\{\left[(d=2, w=10), (d=3, w=10)\right], \left[(d=2, w=5), (d=3, w=5)\right]\}$ 
could belong to the same environment (depth $2 \rightarrow 3$), whereas this would not be allowed in \cref{sec:methods}.
Hence, one environment in this setting corresponds to the union of multiple environments in the setting described at \cref{sec:methods}.
Achieving robustness in this family of environments may be significantly easier due to further averaging, which can mask failures of robustness in some hyperparameter configurations.

\begin{figure}[h]
    \vspace{1cm}
    \centering
    \includegraphics[width=\textwidth]{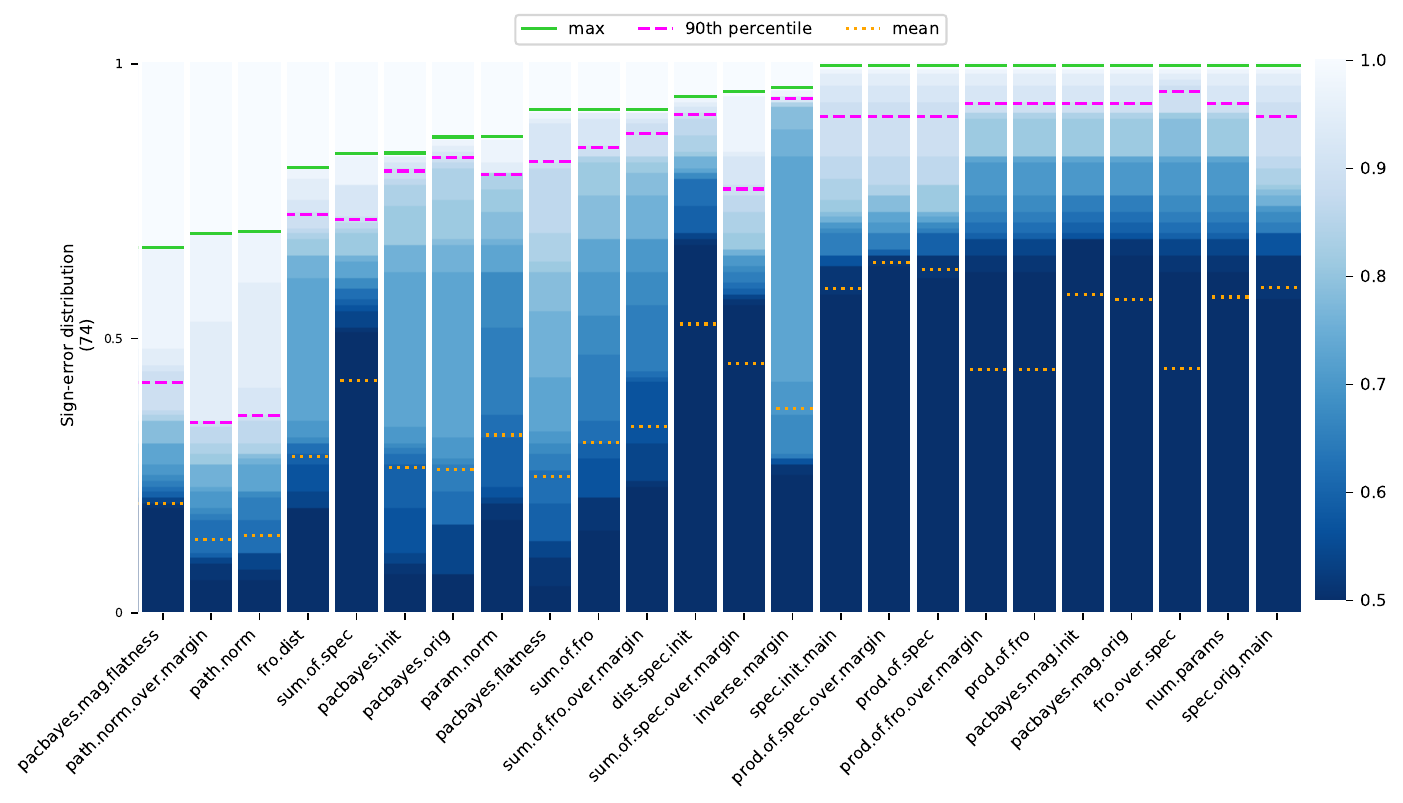}
    \caption{Cumulative distribution of the sign-error across an alternative family of environments where the values of 
    only one pair of variables are fixed and the error is averaged over the rest of variable settings. The number of such 
    environments is shown in parenthesis.
    The measures are ordered based on their robust sign-error (max) across `All' environments.
    A completely \emph{white} bar indicates that the measure is perfectly robust, whereas a \emph{dark blue} bar indicates that it completely fails to be robust.}
    \label{fig:rankingeasy}
\end{figure}

As illustrated in \cref{fig:rankingeasy}, this setting appears to be less challenging than the one described in \cref{sec:methods}. 
In fact, in \cref{fig:ranking_cifar10_svhn_cdf_per_param}, we observe that
all measures achieve a robust sign-error (max value) of 1. However, in \cref{fig:rankingeasy}, we observe that some
measures, such as \texttt{pacbayes.mag.flatness} and \texttt{path.norm.over.margin}, achieve a maximum sign-error much lower than 1.
It is thus apparent that the further averaging that occurs in this setting leads to an easier task and prevents some failures in robustness of being exposed.

We therefore confirm that the set of environments considered in the main text is more challenging and that it constitutes a more relevant setting for studying the robustness of generalization measures.

\clearpage
\section{Methodological comparison to the conditional-independence testing method of \citet{FGM}}\label{app:jiang-ic}

In addition to their Kendall-$\tau$-based $\Psi$ measure, \citet{FGM} propose a measure based on conditional-independence testing
(Section 2.2.3 in \cite{FGM}). This measure attempts to identify the existence of a causal relationship (edge in a postulated causal graph) between a generalization
measure and the generalization gap. While the concept of robustness (that our work builds on) and conditional-independence testing can both
be tied to the causal inference literature, our methods are fundamentally different. Notably, we do not claim, nor seek, to identify
causal relationships (see the discussion in \cref{svmexample}).
Below, we highlight some key differences between our approaches.

It can be tempting to see a similarity in the fact that both methods look at extreme values (min and max).
Our method looks at the maximum sign-error over all environments, which is analogous to taking the max over 
every possible intervention on each hyperparameter. The method of \citet{FGM} considers the minimum normalized conditional mutual 
information ($\mathcal{\hat I}$; Eq. (13) of \cite{FGM}) over all conditioning sets of two hyperparameters (Eq. (14) of \cite{FGM}). However, as described in 
their Eq. (11) and (12), the calculation of $\mathcal{\hat I}$ involves \emph{averaging over all values} of the hyperparameters 
in the conditioning set (i.e., $\sum_{U_\mathcal{S}} p(U_\mathcal{S})$).
Therefore, while they may appear similar, both approaches are fundamentally different in that one averages over multiple values
of the same hyperparameter, while the other (ours) does not.

Furthermore, in the limit where we observe every possible intervention on the HPs, our method
identifies measures that may have a causal relationship to generalization (all causal explanations are necessarily robust in this extreme case).
However, this is not necessarily true for the IC-based method of \citet{FGM}.
The reason is that their conditioning sets are of size 2, which may leave open confounding paths in the graph if more than 2 hyperparameters
act as confounders, resulting in non-causal mutual information.
This means that the method is not guaranteed to detect a causal edge from the generalization measure to generalization gap unless they condition on all hyperparameters.
However, if they were to condition on all hyperparameters, their conditional mutual information would collapse to zero, 
suggesting that there is no causal edge.

\end{document}